\pdfoutput=1
\documentclass[11pt]{article}

\usepackage[preprint]{acl}
\usepackage{times}
\usepackage{latexsym}
\usepackage[T1]{fontenc}
\usepackage[utf8]{inputenc}
\usepackage{xcolor} 
\usepackage{amsmath} 
\usepackage{tcolorbox}
\usepackage{colortbl}
\usepackage{microtype}
\usepackage{inconsolata}
\usepackage{graphicx}
\usepackage{subcaption}
\usepackage{booktabs}
\usepackage{multirow}
\definecolor{myblue}{RGB}{150, 150, 230}
\usepackage{amsmath,amsfonts,bm}

\def\eqref#1{equation~\ref{#1}}
\def\1{\bm{1}}

\def\vh{{\bm{h}}}

\def\vv{{\bm{v}}}

\def\mA{{\bm{A}}}

\def\mD{{\bm{D}}}

\def\mH{{\bm{H}}}
\def\mI{{\bm{I}}}

\def\mT{{\bm{T}}}

\def\mV{{\bm{V}}}

\DeclareMathAlphabet{\mathsfit}{\encodingdefault}{\sfdefault}{m}{sl}
\SetMathAlphabet{\mathsfit}{bold}{\encodingdefault}{\sfdefault}{bx}{n}

\def\gL{{\mathcal{L}}}

\newcommand{\R}{\mathbb{R}}

\DeclareMathOperator*{\llm}{LLM}

\usepackage{tablefootnote}
\usepackage{hyperref}

\title{
Rethinking Visual Dependency in Long-Context Reasoning \\for Large Vision-Language Models
}
\def\spaces{~~~~~}
\author{Yucheng Zhou$^{1}$\spaces{}
        Zhi Rao$^{2}$\spaces{}
        Jun Wan$^{2}$\spaces{}
        Jianbing Shen$^{1}$\thanks{~Corresponding author.}\\
        $^{1}$ SKL-IOTSC, CIS, University of Macau \\
        $^{2}$ Macau University of Science and Technology \\
        {\tt yucheng.zhou@connect.um.edu.mo, jianbingshen@um.edu.mo}
    }

\begin{document}
\maketitle

\begin{abstract}
Large Vision-Language Models (LVLMs) excel in cross-model tasks but experience performance declines in long-context reasoning due to overreliance on textual information and reduced visual dependency. In this study, we empirically analyze LVLMs in long-context reasoning, revealing that increased context length leads to a higher dependence on language at the expense of visual dependency. To address this issue, we propose a novel training-free context pruning method that selectively removes less critical textual information. Our approach enhances visual dependency and reduces textual noise, thereby improving LVLM performance in long-context reasoning. We validate our method by constructing a long-context dataset, demonstrating its effectiveness across various LVLMs. Moreover, we compare our method with token-pruning methods, demonstrating superior performance, and further analysis confirms the robustness of our method.
\end{abstract}

\section{Introduction}
Large Vision-Language Models (LVLMs) have achieved remarkable success across a range of cross-modal understanding and reasoning~\citep{InstructBLIP,Qwen-VL,MiniGPT4}. However, when LVLMs are applied to long dialogues or complex reasoning tasks, their performance tends to degrade. For instance, in our pilot experiments, LLaVA~\citep{llava} exhibits up to a 17\% decline in performance under long-context reasoning. This degradation shows that the model does not accurately understand the visual input, suggesting that LVLMs are overly dependent on textual priors, diminishing the visual dependency.

Recently, Large Language Models (LLMs) have significantly enhanced their performance in long-context reasoning through contextual position encoding and optimized attention strategies~\citep{LongRoPE,LongLM}. These methods enable LLMs to retain and effectively utilize more information over long-context interactions, thereby maintaining contextual coherence and improving response accuracy. In contrast, LVLMs struggle to leverage visual information appropriately when processing long contexts. Therefore, LVLMs are required to integrate visual and textual information under long-context reasoning accurately.

In this work, we conduct extensive analysis for LVLMs under long-context reasoning. Empirical results indicate a significant performance drop in LVLM as context length increases, while the opposite is true without visual information. It demonstrates that the model could rely more on the language priors with longer contexts, reducing its visual dependency. Moreover, our analysis further reveals that as the proportion of the target object within the image increases, the model's visual dependency exhibits greater stability. By analyzing the attention weights within the model, we observe more cross-modal interaction in the shallow layers, followed by predominantly textual interaction in the deeper layers.

To improve the visual dependency in long-context reasoning, we propose a novel training-free context pruning method for LVLMs, which strategically reduces the interference of less critical textual information while preserving visual information. Our approach leverages the model's internal attention mechanisms to identify and eliminate the trivial textual tokens based on their aggregated attention weights across multiple attention heads. By systematically pruning these less significant tokens, our method effectively diminishes textual noise and encourages more attention to visual information, enhancing visual dependency under long-context reasoning. In addition, we provide a theoretical foundation demonstrating that context pruning leads to more concentrated and stable attention distributions, which in turn fosters a stronger visual dependency on visual inputs. 
This theoretical demonstration is further validated through a detailed analysis of information flow.

To comprehensively evaluate the performance of LVLMs in long-context reasoning, we construct a long-context dataset based on the SVIT dataset~\citep{SVIT}. Experimental results demonstrate the effectiveness of our approach across various LVLMs. 
Moreover, we compare our method with other token pruning approaches on the MMDU dataset, achieving superior performance. Our method demonstrates significant improvements across different model sizes.
Additionally, we further verify the robustness of our method by comparing different token pruning strategies, as well as various pruning layers and rates. The characteristics of our pruning method are further analyzed through part-of-speech observations of the pruned tokens. Lastly, we conduct an exploration of the scaling laws about the relationship between token pruning rates and context length.

In this study, our contributions are as below:
\begin{itemize}
    \item \vspace{-2mm}We conduct an empirical analysis of LVLMs in long-context reasoning, exploring the causes of performance decline and the factors influencing visual dependency.
    \item \vspace{-2mm}We propose a novel train-free context pruning method that reduces less relevant textual information, improving visual dependency and performance in long-context reasoning.
    \item \vspace{-2mm}We perform theoretical and information flow analysis demonstrating context pruning leads to more concentrated and stable attention distributions, which enhances visual dependency.
    \item \vspace{-2mm}We construct a long-context dataset based on SVIT and validate our approach through comprehensive experiments, also analyzing pruning strategies and exploring scaling laws between pruning rates and context length.
\end{itemize}

\section{Background and Notation}
LVLMs include a vision encoder and a LLM, and their input consists of images and text, which are processed through cross-modal interaction. 
The vision encoder maps the input image $\mI$ to $n$ low-dimensional feature vectors $\mV_I = \vv_i \in \R^d, i \in \{1, 2, \cdots, n\}$. 
While the text embedding layer transforms the input text sequence $\mT = \{t_1, t_2, ..., t_m\}$ into a sequence of embeddings $\mH_T = \{\vh_1, \vh_2, ..., \vh_m\}$, where each $\vh_i \in \R^d$ represents the embedding of the $i$-th token. 
The visual features $\mV_I$ and the text embeddings $\mH_T$ are fed into the LLM, predicting subsequent tokens in the text sequence by cross-modal interaction, i.e.,
\begin{align}
p(t_{m+1} | \mT, \mI) = \llm(\mH_T, \mV_I).
\end{align}
The visual dependency $\mD_{\text{vis}}(k)$ measures the extent to which the language model relies on visual features $\mV_I$ when generating the $k$-th token. Specifically, the attention weights $\alpha_{k,i}$ at position $k$ are distributed among all available inputs, i.e., textual embeddings $\mH_T$ and visual features $\mV_I$. 
As the context length $m$ increases, the attention weights are normalized over a larger set of tokens, resulting in smaller visual attention weights, i.e.,
\begin{align}
\mathcal{D}_{\text{vis}}(k) = \sum_{i=1}^{n} \alpha_{k,i},
\end{align}
where $\alpha_{k,i}$ is the attention weight from position $k$ to visual feature vector $\vv_i$. Since the attention mechanism normalizes the weights over all input tokens, increasing textual tokens $m$ can result in smaller attention weights $\alpha_{k,i}$ for the visual features. Consequently, the impact of $\mV_I$ on the model's predictions diminishes as the textual context grows, which presents a decrease in visual dependency.

\vspace{-1.5mm}
\paragraph{Related Work (Summarized). }Recent advancements in large vision-language models (LVLMs) focus on enhancing cross-modal understanding by integrating visual and textual inputs for complex tasks~\citep{VisionLLM,MiniGPTv2,ShareGPT4V}. Models like LLaVA~\citep{llava} and LLaVA-NeXT~\citep{LLaVANeXT} improve performance in multimodal reasoning and OCR, while approaches like DeepSeek-VL~\citep{DeepSeekVL} address challenges of noisy inputs. In long-context reasoning, techniques like Contextual Position Encoding (CoPE)~\citep{Contextual} and sliding windows~\citep{LongContext} extend context handling, with methods such as ThoT~\citep{ThoT} and LLMLingua~\citep{LLMLingua} improving reasoning across fragmented contexts. The full version can be found in Appendix~\ref{app:related}.

\section{Analysis on Visual Dependency of LVLMs in Long-Context Reasoning}

\subsection{Experimental Setting.}\label{sec:setting}
To analyze the visual dependency of LVLMs in long-context reasoning, we construct a long-context dataset\footnote{The dataset and code will be made publicly available after the paper's acceptance to encourage community adoption.} based on the SVIT dataset~\citep{SVIT}. SVIT dataset comprises images with multiple types of annotations, including conversation, complex reasoning, referring QAs, and detail description. These annotations can be concatenated to form long text annotations for images. We retain a total of 15,295 samples through random sampling and filtering. The samples are proportionally distributed into groups of 0.5k, 1k, 1.5k, 2k, and 2.5k based on text length. Each sample is processed in a question-answer format, with filtering to ensure that the objects in the answers do not appear in the context provided for the questions. For the analysis, we used the following LVLMs with 7B parameters: DeepSeek-VL~\citep{DeepSeekVL}, LLaVA~\citep{llava}, LLaVA-Next~\citep{LLaVANeXT}, ViP-LLaVA~\citep{ViPLLaVA}, LLaVA-Next-Video~\citep{LLaVANeXTVideo}, and Video-LLaMA2~\citep{VideoLLaMA2}. The performance metrics on this dataset refer to accuracy, measured by the correctness of inference results.

\subsection{Does Context Length Affect Visual Dependency?}
\begin{figure}[t]
    \centering
    \includegraphics[width=\linewidth]{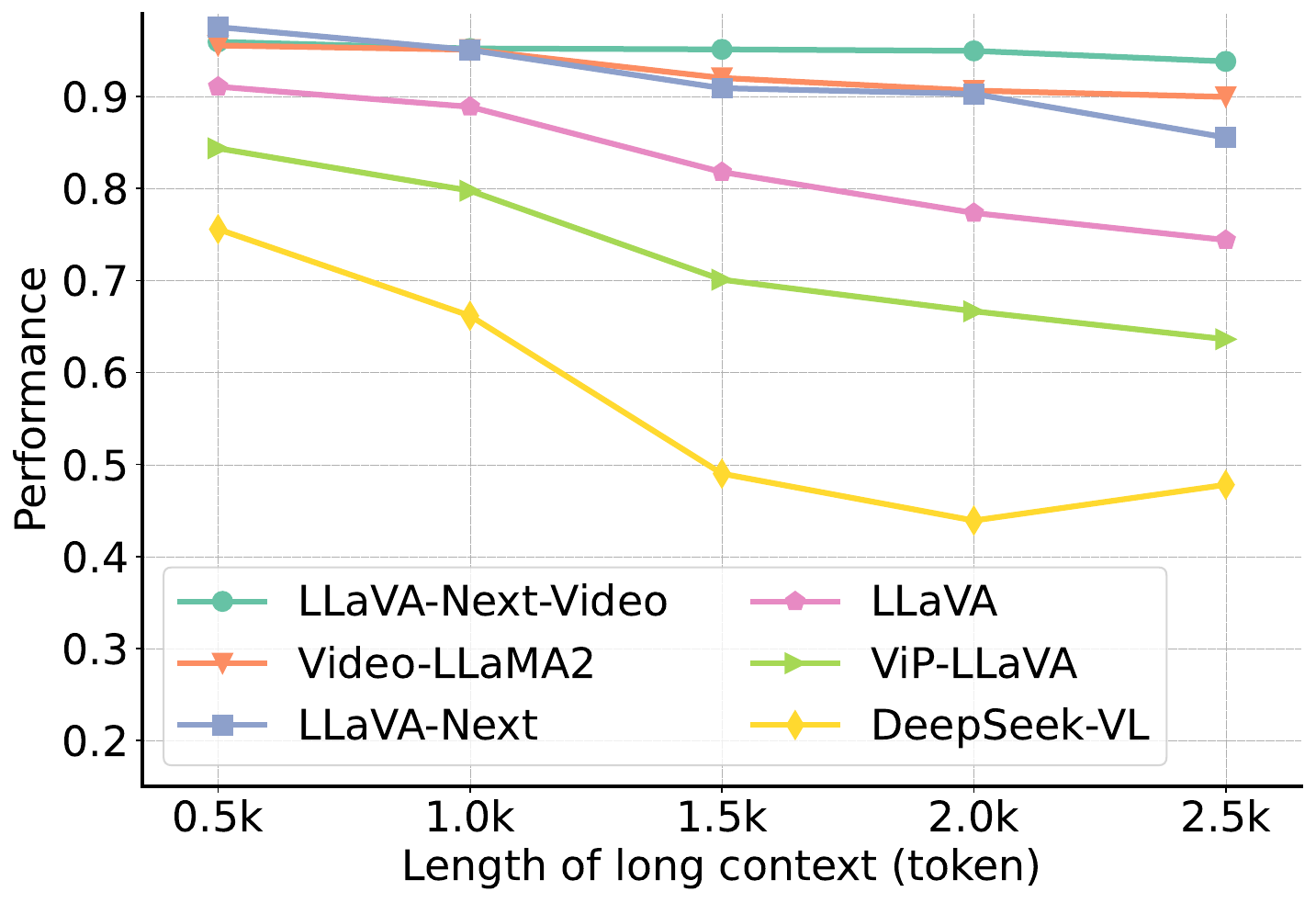}
    \vspace{-8mm}
    \caption{\small Performance of LVLMs on long-context reasoning.}
    \label{fig:visual_long_context}
    \vspace{-4mm}
\end{figure}

To investigate whether context length affects the visual dependency of LVLMs, we conducted experiments on various LVLMs with progressively long contexts. As shown in Figure~\ref{fig:visual_long_context}, we observe that the performance of various LVLMs declines as the context length increases. DeepSeek-VL experiences an approximate 28\% drop in performance. 
Moreover, LVLMs trained on video data, i.e., LLaVA-NeXT-Video and Video-LLaMA2, exhibit greater robustness on long text, with relatively smaller reductions in performance. 
% However, their performance also diminishes with increased context length. 
It indicates as the text length increases, the dependency of LVLMs on visual input decreases, leading to a decline in performance.

\paragraph{Impact of Language Priors.}
\begin{figure}[t]
    \centering
    \includegraphics[width=\linewidth]{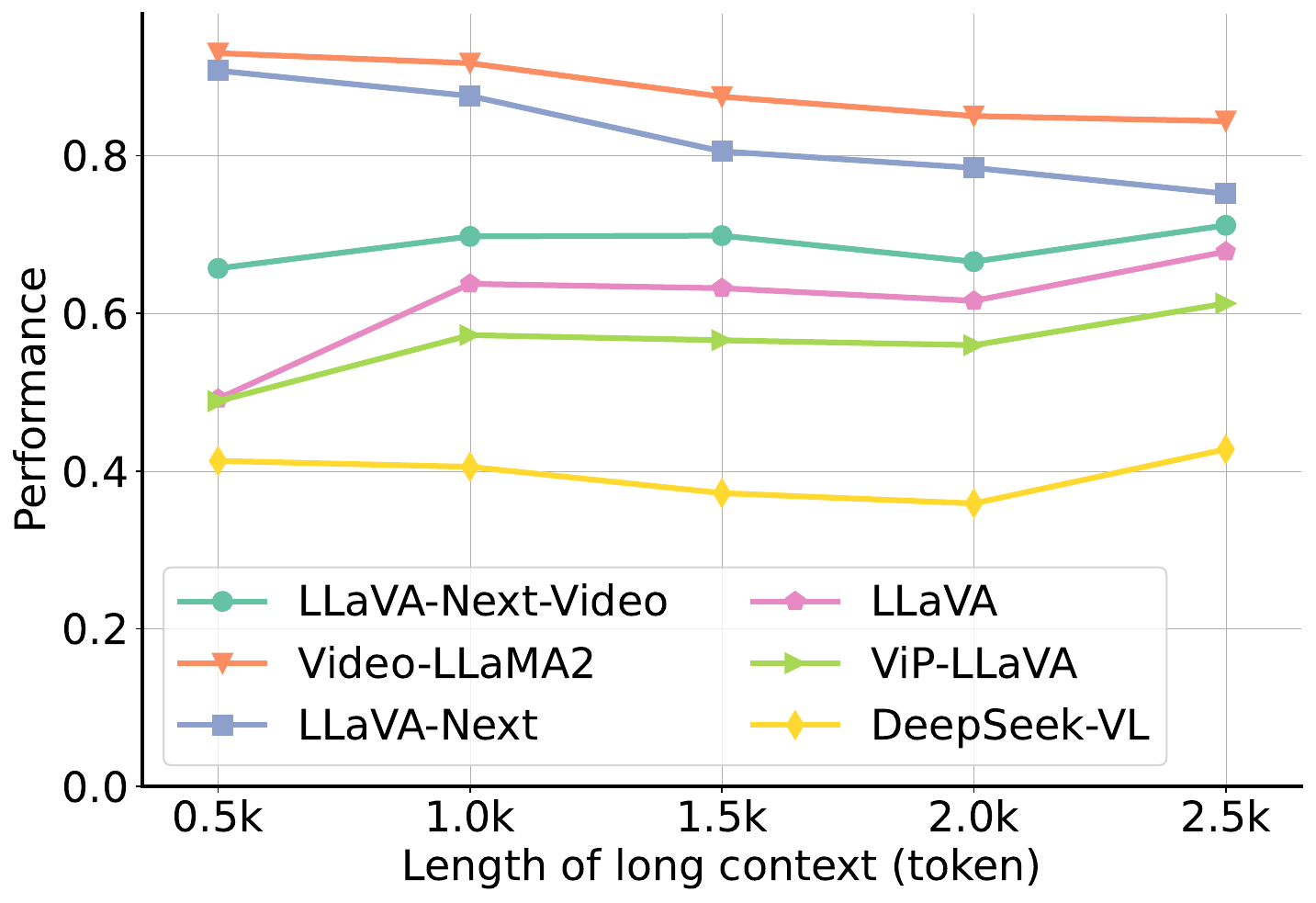}
    \vspace{-8mm}
    \caption{\small Performance of LVLMs on text-only long contexts.}
    \label{fig:text_long_context}
    \vspace{-4mm}
\end{figure}
In addition to relying on visual input to infer the correct answers, LVLMs can also leverage language priors to make predictions. For instance, objects like ``desk'' and ``chair'' tend to co-occur frequently, allowing the model to use this co-occurrence probability as a cue. To investigate the impact of language priors, we replaced the images in our samples with either solid white or black images (to maintain identical token lengths) and fed these modified inputs into the LVLMs.
As shown in Figure~\ref{fig:text_long_context}, most LVLMs perform better as the context length increases in the presence of strong language priors. This improvement occurs because longer contexts provide more details, making it easier for the model to infer the presence of related objects based on prior knowledge. However, LLaVA-Next maintains a relatively high performance overall. This can be attributed to the instruction tuning datasets of LLaVA-Next, which include same images from SVIT dataset. Consequently, LLaVA-Next has been trained to recognize and infer visual content based on contextual cues, even without images.

\subsection{Consistency Between Vision and Language Priors}
\begin{figure}[t]
    \centering
    \includegraphics[width=0.9\linewidth]{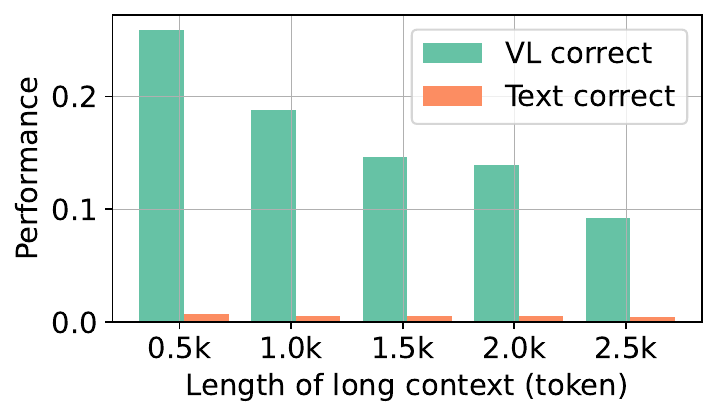}
    \vspace{-4mm}
    \caption{\small Consistency between vision and language priors.}
    \label{fig:consistency}
    \vspace{-5mm}
\end{figure}
We analyze the consistency between vision and language priors by comparing prediction performance with and without visual input across varying context lengths. Figure \ref{fig:consistency} illustrates the average accuracy of six models, showing that samples correctly answered with language priors alone remained accurate even when visual cues were added, highlighting the robustness of language reasoning. The detailed analysis can be found in the Appendix~\ref{app:consistency}.

\subsection{How Does Visual Content Affect Visual Dependency?}
\begin{figure}[t]
    \centering
    \includegraphics[width=0.9\linewidth]{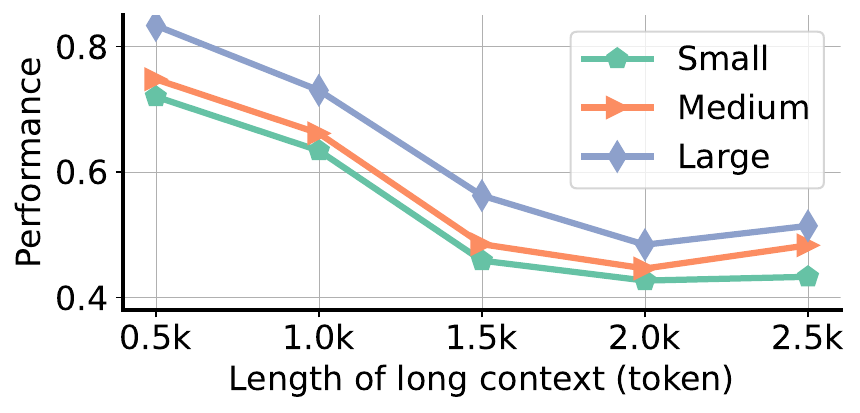}
    \vspace{-4mm}
    \caption{\small The relationship between the proportion of target visual content in the image and performance.}
    \label{fig:visual_content}
    \vspace{-4mm}
\end{figure}
To analyze the impact of varying proportions of target visual content on model performance, we split samples based on the bounding box sizes of the target objects into three distinct groups: ``Large'', ``Medium'' and ``Small''. Figure~\ref{fig:visual_content} shows each category's performance across different context lengths.
From the figure, the model performance generally decreases as the context length increases. This trend is observed across all three groups. In particular, the group with ``Large'' bounding box values shows superior performance compared to the ``Medium'' and ``Small'' groups, which demonstrates that larger target objects (i.e., more significant visual content) allow the model to capture and utilize visual dependencies better.

\subsection{Attention Weight Distribution Across Vision and Language Information}
\begin{figure}[t]
    \centering
    \includegraphics[width=\linewidth]{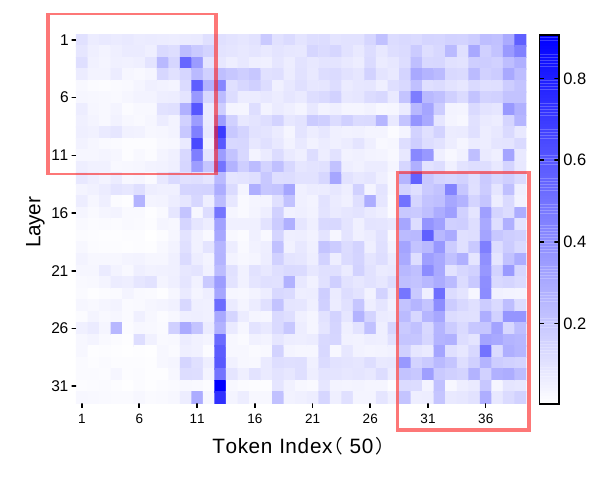}
    \vspace{-9mm}
    \caption{\small The distribution of attention weights between different layers of LVLMs.}
    \vspace{-5mm}
    \label{fig:attention}
\end{figure}
To further investigate the model's visual dependency, we analyze the distribution of attention weights in LVLMs from the perspective of cross-modal interactions. As shown in Figure~\ref{fig:attention}, the attention distribution across different layers of the LLM backbone is visualized with 50 tokens per unit on the x-axis. The sequence is divided into two parts: the left side represents attention weights associated with visual information, and the right side corresponds to textual information. In the shallow layers (highlighted by the red box on the left), attention is heavily concentrated on the visual tokens, indicating that these layers are primarily responsible for processing and integrating visual inputs. Meanwhile, we observe a shift in attention towards the textual tokens (highlighted on the right), suggesting that deep layers focus more on language processing while still considering the visual context.

\section{Context Pruning to Improve Visual Dependency}
\begin{figure*}[t]
    \centering
    \includegraphics[width=\linewidth]{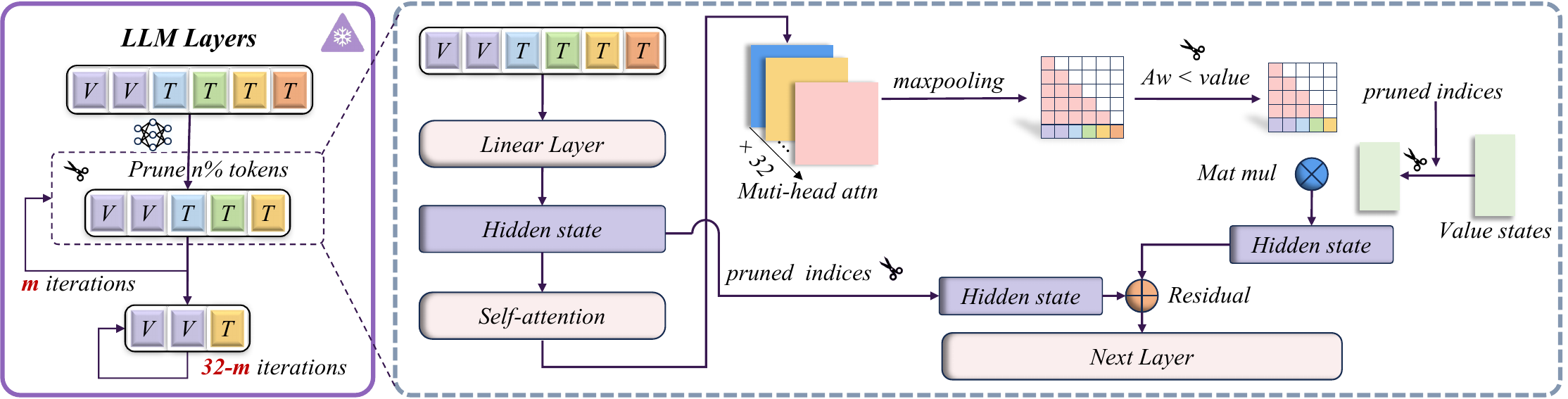}
    \vspace{-6mm}
    \caption{\small Overview of our method. \textbf{Left:} token pruning across multiple iterations; \textbf{Right:} pruning tokens in each LLM layer.}
    \label{fig:method}
    \vspace{-5mm}
\end{figure*}

For long-context reasoning, the model's attention weights become more distributed as the input length increases. It leads the model to rely more on textual priors, diminishing its dependence on visual content. As shown in Figure~\ref{fig:method}, we propose a training-free context pruning method designed to enhance the model's visual dependency under long-context reasoning. By selectively pruning tokens in the textual input, we aim to reduce irrelevant information while retaining critical contextual cues. Moreover, we theoretically demonstrate the effectiveness and reliability of our method.

\subsection{Context Pruning}
Our method involves pruning a percentage $n\%$ of the least important tokens in the textual input based on their attention weights in the Transformer layers. Meanwhile, we ensure that visual tokens remain unpruned, allowing the model to retain complete access to visual information. The pruning is performed by first calculating the attention weights in each of the first $m$ layers of the Transformer and then ranking these weights to determine which tokens to prune. For tokens in multi-head attention, we perform max-pooling overheads to obtain a unified importance score for each token. We then prune $n\%$ tokens with the lowest attention scores.

Specifically, given a text $T = \{t_1, t_2, \cdots, t_n\}$ and an image $V = \{v_1, v_2, \cdots, v_k\}$, we compute the attention weights $A^{(l)}_{ij}$ between each pair of tokens $t_i$ and $t_j$ in $l$-th Transformer layer, i.e.,
\begin{align}
A^{(l)}_{ij} = \frac{\exp\left(\frac{Q^{(l)}_i \cdot K^{(l)}_j}{\sqrt{d_k}}\right)}{\sum_{k=1}^{n} \exp\left(\frac{Q^{(l)}_i \cdot K^{(l)}_k}{\sqrt{d_k}}\right)}
\end{align}
where $Q^{(l)}_i$ and $K^{(l)}_j$ are the query and key representations of tokens $t_i$ and $t_j$, respectively, in layer $l$, and the attention weight $A^{(l)}_{ij}$ measures how much token $t_i$ attends to token $t_j$ in that layer. $\frac{1}{\sqrt{d_k}}$ is scaling factor, and $d_k$ the dimensionality of $K^{(l)}_j$.

For multi-head attention, we aggregate the attention weights across all heads by applying max-pooling. The score for token $t_i$ is obtained by taking the maximum across all heads, i.e., 
\begin{align}
\hat{A}^{(l)}_i = \max_{h} \left( \sum_{j=1}^{n} A^{(l,h)}_{ij} \right)
\end{align}
where $A^{(l,h)}_{ij}$ represent the attention weights for head $h$ in layer $l$. Once the scores $\hat{A}^{(l)}_i$ are calculated for all tokens, we sort these scores in ascending order and prune the bottom $n\%$ of tokens with the smallest scores. Only textual tokens $T$ are considered for pruning, while the visual tokens $V$ remain unpruned throughout the process. This ensures that visual information is fully preserved.

\subsection{Theoretical Demonstration}
Our method aims to improve LVLMs by selectively pruning text tokens, thereby reducing text noise and forcing the model to depend more on visual cues. We provide a theoretical demonstration to explain how pruning based on attention weights can lead to more stable and accurate attention distributions, and how this pruning enhances visual dependency.

\begin{tcolorbox}[colback=gray!15, colframe=black, left=1mm, right=1mm, top=1mm, bottom=1mm]
\paragraph{Proposition 1:} \textit{After pruning low-importance tokens, the attention distribution becomes more concentrated on important tokens.}
\end{tcolorbox}

\paragraph{Proof:}
Let $X = [x_1, x_2, \dots, x_T]$ be the input sequence of tokens at the $l$-th layer, where $T$ is the total number of tokens. The attention weight matrix $A^{(l)}$ contains the attention weights $A^{(l)}_{ij}$, where $A^{(l)}_{ij}$ denotes the attention weight of token $x_i$ over token $x_j$. The attention distribution for token $x_i$ is denoted as $A_i^{(l)} = [A^{(l)}_{i1}, A^{(l)}_{i2}, \cdots, A^{(l)}_{iT}]$, and can be seen as a probability distribution with $\sum_{j=1}^{T} A^{(l)}_{ij} = 1$. The concentration of attention can be measured using entropy, i.e.,
\begin{align}
H(A_i^{(l)}) = - \sum_{j=1}^{T} A^{(l)}_{ij} \log A^{(l)}_{ij}
\end{align}
where smaller entropy indicates higher attention concentration, meaning the model focuses more on specific tokens. 
Consider pruning the lowest $n\%$ of tokens based on attention weights. Let the remaining tokens form the pruned sequence $\bar{X} = [\bar{x}_1, \bar{x}_2, \cdots, \bar{x}_{\bar{T}}]$, where $\bar{T} = (1 - n\%) \times T$. After pruning, the attention distribution of token $x_i$ is updated to $\bar{A}^{(l)}_{i}$, and the entropy are defined as:
\begin{align}
H(\bar{A}^{(l)}_{i}) = - \sum_{j=1}^{\bar{T}} \bar{A}^{(l)}_{ij} \log \bar{A}^{(l)}_{ij}
\end{align}

Since the low-weight tokens have been pruned, remaining weights are concentrated on more important tokens, leading to entropy reduction:
\begin{align}
H(\bar{A}^{(l)}_{i}) \leq H(A^{(l)}_i)
\end{align}

Thus, pruning reduces the entropy of attention distributions, making the model's focus more stable and accurate.
To quantify the attention stability, we can calculate the variance of the attention weights before and after pruning for token $x_i$, i.e.,
\begin{align}
\sigma^2(A^{(l)}_i) &= \frac{1}{T} \sum_{j=1}^{T} \left( A^{(l)}_{ij} - \mu(A^{(l)}_i) \right)^2 \notag\\
\sigma^2(\bar{A}^{(l)}_{i}) &= \frac{1}{\bar{T}} \sum_{j=1}^{\bar{T}} \left( \bar{A}^{(l)}_{i} - \mu(\bar{A}^{(l)}_{i}) \right)^2
\end{align}
where $\sigma^2(A^{(l)}_i)$ and $\sigma^2(\bar{A}^{(l)}_{i})$ are the variance of the attention weights before and after pruning, respectively. Since pruning low-weight tokens increases the concentration of attention on the remaining tokens, variance of attention distribution increases:
\begin{align}
\sigma^2(\bar{A}^{(l)}_{i}) \geq \sigma^2(A^{(l)}_i)
\end{align}
where the increase in variance indicates the remaining attention is more focused on important tokens.

\begin{tcolorbox}[colback=gray!15, colframe=black, left=1mm, right=1mm, top=1mm, bottom=1mm]
\paragraph{Proposition 2:} \textit{As textual tokens are pruned, the model becomes more dependent on visual information for reasoning.}
\end{tcolorbox}

\paragraph{Proof:}
Consider a LVLM processes both visual input $V$ and textual input $T$ to produce an output $O$, such that $O = f(V, T)$. The model's dependence on each modality can be quantified through its attention mechanism and the sensitivity of the output to each input.
Let $A_V$ and $A_T$ represent the attention distributions over visual and textual inputs, satisfying the normalization condition:
\begin{align}
\sum_{i} A_{V,i} + \sum_{j} A_{T,j} = 1,
\end{align}
where $A_{V,i}$ and $A_{T,j}$ are the attention weights assigned to the $i$-th visual feature and the $j$-th textual token.
Define $\alpha$ and $\beta$ as the total attention weights allocated to visual and textual inputs, respectively:
\begin{align}
\alpha = \sum_{i} A_{V,i}, \quad \beta = \sum_{j} A_{T,j}.
\end{align}

The model's sensitivity to each modality can be linked to these attention weights. Specifically, the sensitivity of the output $O$ to the visual input $V$ is proportional to $\alpha$, and the sensitivity to the textual input $T$ is proportional to $\beta$:
\begin{align}
\left\| \frac{\partial O}{\partial V} \right\| \propto \alpha, \quad \left\| \frac{\partial O}{\partial T} \right\| \propto \beta.
\end{align}

When textual tokens are pruned, the textual input reduces from $T$ to a pruned version $\bar{T}$, containing fewer tokens. The attention weights over textual inputs decrease, leading to a reduced $\beta$:
\begin{align}
\beta' = \sum_{j} \bar{A}_{T,j} < \beta,
\end{align}
where $\bar{A}_{T,j}$ are the attention weights assigned to the remaining textual tokens after pruning.
Since the total attention must still sum to 1, the attention allocated to visual inputs increases:
\begin{align}
\alpha' = \sum_{i} \bar{A}_{V,i} = 1 - \beta' > 1 - \beta = \alpha.
\end{align}
where $\alpha' > \alpha$ indicates increased dependence on visual inputs.
This increase in $\alpha$ directly translates to greater sensitivity of output $O$ to visual input $V$:
\begin{align}
\left\| \frac{\partial O}{\partial V} \right\|' \propto \alpha' > \alpha \propto \left\| \frac{\partial O}{\partial V} \right\|.
\end{align}
Similarly, sensitivity to textual input decreases:
\begin{align}
\left\| \frac{\partial O}{\partial \bar{T}} \right\|' \propto \beta' < \beta \propto \left\| \frac{\partial O}{\partial T} \right\|.
\end{align}
Therefore, as textual tokens are pruned, the model reallocates its attention from textual to visual inputs, resulting in increased $\alpha'$ and decreased $\beta'$. This shift enhances the model's dependency on visual information for reasoning, as evidenced by both the attention weights and sensitivity measures.

\subsection{Information Flow Analysis}
Following the approaches \cite{VICL,wordsanchor}, we further validate the effectiveness of our method through information flow. The saliency score for each element in the attention matrix is computed using an adapted version of the Taylor expansion method \cite{MichelLN19}, i.e.,
\begin{align}
I_l = \sum_h \left|\mA_{h, l}^{\top} \odot \frac{\partial \gL(x)}{\partial \mA_{h, l}}\right|,
\end{align}
where $h$ and $l$ denote distinct attention heads and Transformer layers, respectively, and $\gL(x)$ represents the loss function. We propose six significance scores based on information flow, i.e.,
\begin{align}
S_{vr} &= \frac{\sum_{(i,j) \in C_{vr}} I_l(i,j)}{|\{(v, r_k) : k \in [1, R]\}|}, \\
S_{vc} &= \frac{\sum_{(i,j) \in C_{vc}} I_l(i,j)}{|\{(v, c_k) : k \in [1, C]\}|}, \\
S_{rt} &= \frac{\sum_{(i,j) \in C_{rt}} I_l(i,j)}{|\{(r_k, t) : k \in [1, R]\}|}, \\
S_{ct} &= \frac{\sum_{(i,j) \in C_{ct}} I_l(i,j)}{|\{(c_k, t) : k \in [1, C]\}|}, \\
S_{vt} &= \frac{\sum_{(i,j) \in C_{vt}} I_l(i,j)}{|\{(v, t)\}|}, \\
S_{ww} &= \frac{\sum_{(i,j)\in C_{ww}} I_l(i,j)}{|C_{ww}|},
\end{align}
where $v, r_k, c_k, t$, $R$, and $C$ represent the visual part, preserved tokens, pruned tokens, the target position, the total number of preserved tokens, and the total number of pruned tokens, respectively. 
\begin{table*}[!t]\small
\centering
\resizebox{\textwidth}{!}{
\begin{tabular}{lllllll}
\toprule
\multirow{2}{*}{\bf Model} & \multirow{2}{*}{\bf Method} & \multicolumn{5}{c}{\bf Context Length (token)} \\
\cmidrule(lr){3-7}
 & & \multicolumn{1}{c}{\bf 0.5k} & \multicolumn{1}{c}{\bf 1.0k} & \multicolumn{1}{c}{\bf 1.5k} & \multicolumn{1}{c}{\bf 2.0k} & \multicolumn{1}{c}{\bf 2.5k} \\
\midrule
\multirow{2}{*}{DeepSeek-VL~\citep{DeepSeekVL}} 
& Baseline & 75.60 & 66.20 & 49.00 & 43.90 & 47.80 \\
& \cellcolor{gray!15} Ours & \cellcolor{gray!15}78.85\textcolor{blue}{(3.25)} & \cellcolor{gray!15}71.98\textcolor{blue}{(5.78)} & \cellcolor{gray!15}67.87\textcolor{blue}{(18.87)} & \cellcolor{gray!15}66.33\textcolor{blue}{(22.43)} & \cellcolor{gray!15}62.63\textcolor{blue}{(14.83)} \\
\midrule
\multirow{2}{*}{ViP-LLaVA~\citep{ViPLLaVA}} 
& Baseline & 84.40 & 79.80 & 70.10 & 66.70 & 63.60 \\
& \cellcolor{gray!15} Ours & \cellcolor{gray!15}91.57\textcolor{blue}{(7.17)} & \cellcolor{gray!15}91.50\textcolor{blue}{(11.70)} & \cellcolor{gray!15}87.28\textcolor{blue}{(17.18)} & \cellcolor{gray!15}83.79\textcolor{blue}{(17.09)} & \cellcolor{gray!15}75.74\textcolor{blue}{(12.14)} \\
\midrule
\multirow{2}{*}{LLaVA~\citep{llava}} 
& Baseline & 91.10 & 88.90 & 81.80 & 77.30 & 74.40 \\
& \cellcolor{gray!15} Ours & \cellcolor{gray!15}95.91\textcolor{blue}{(4.81)} & \cellcolor{gray!15}95.36\textcolor{blue}{(6.46)} & \cellcolor{gray!15}93.92\textcolor{blue}{(12.12)} & \cellcolor{gray!15}91.40\textcolor{blue}{(14.10)} & \cellcolor{gray!15}87.05\textcolor{blue}{(12.65)} \\
\midrule
\multirow{2}{*}{LLaVA-Next~\citep{LLaVANeXT}} 
& Baseline & 97.50 & 95.10 & 90.90 & 90.30 & 85.60 \\
& \cellcolor{gray!15} Ours & \cellcolor{gray!15}97.61\textcolor{blue}{(0.11)} & \cellcolor{gray!15}95.62\textcolor{blue}{(0.52)} & \cellcolor{gray!15}91.47\textcolor{blue}{(0.57)} & \cellcolor{gray!15}90.55\textcolor{blue}{(0.25)} & \cellcolor{gray!15}86.60\textcolor{blue}{(1.00)} \\
\midrule
\multirow{2}{*}{Video-LLaMA2~\citep{VideoLLaMA2}} 
& Baseline & 95.60 & 95.10 & 92.00 & 90.70 & 90.00 \\
& \cellcolor{gray!15} Ours & \cellcolor{gray!15}95.78\textcolor{blue}{(0.18)} & \cellcolor{gray!15}97.42\textcolor{blue}{(2.32)} & \cellcolor{gray!15}97.12\textcolor{blue}{(5.12)} & \cellcolor{gray!15}96.70\textcolor{blue}{(6.00)} & \cellcolor{gray!15}96.31\textcolor{blue}{(6.31)} \\
\midrule
\multirow{2}{*}{LLaVA-Next-Video~\citep{LLaVANeXTVideo}} 
& Baseline & 95.90 & 95.20 & 95.10 & 95.00 & 93.80 \\
& \cellcolor{gray!15} Ours & \cellcolor{gray!15}96.08\textcolor{blue}{(0.18)} & \cellcolor{gray!15}95.52\textcolor{blue}{(0.32)} & \cellcolor{gray!15}95.59\textcolor{blue}{(0.49)} & \cellcolor{gray!15}95.13\textcolor{blue}{(0.13)} & \cellcolor{gray!15}94.12\textcolor{blue}{(0.32)} \\
\bottomrule
\end{tabular}}
\vspace{-3mm}
\caption{\small Performance comparison of our method and baselines across various LVLMs. The \textcolor{blue}{blue} numbers in parentheses indicate the performance gain achieved by our method compared to the baselines.}
\label{tab:result}
\vspace{-5mm}
\end{table*}
\begin{figure}[t]
    \centering
    \vspace{-3mm}
    \includegraphics[width=1\linewidth]{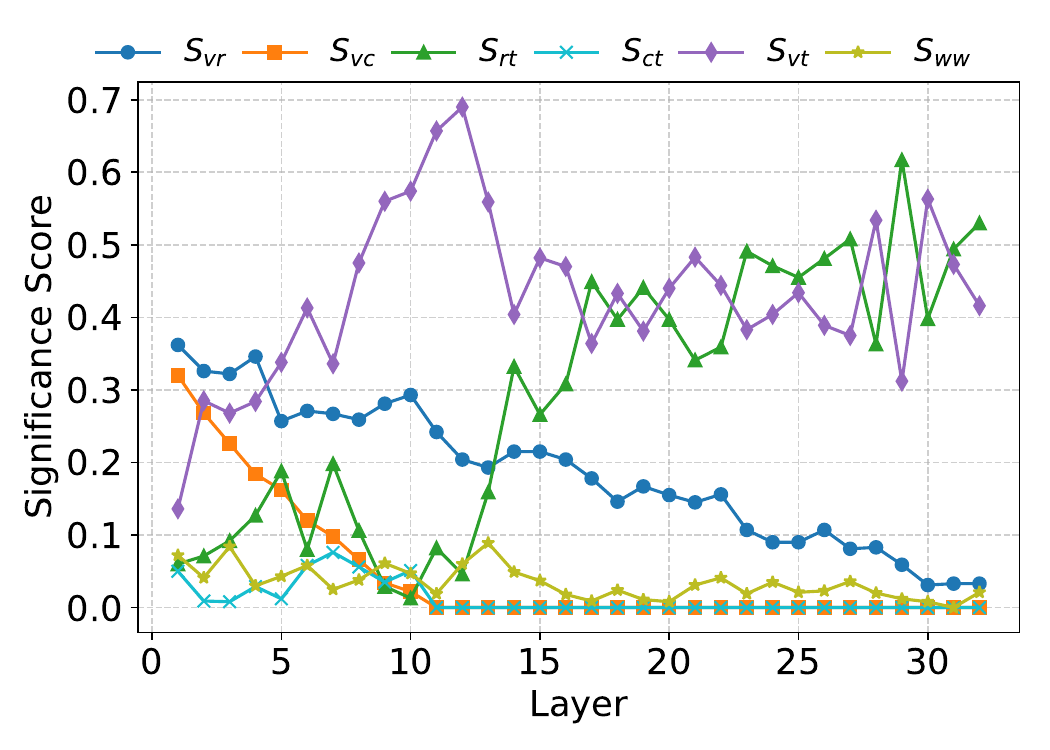}
    \vspace{-9mm}
    \caption{Information flow results.}
    \label{fig:flow}
    \vspace{-2mm}
\end{figure}
\begin{table}[!t]\small
\centering
\setlength{\tabcolsep}{3pt}
\resizebox{\linewidth}{!}{
\begin{tabular}{lcccc}
\toprule
\textbf{Method} & \textbf{LLaVA-Next-Video} & \textbf{Video-LLaMA2} & \textbf{LLaVA-Next} \\
\midrule
Baseline & 44.0~~~~~~~~~~ & 43.7~~~~~~~~~~ & 43.9~~~~~~~~~~ \\
\rowcolor{gray!15}Ours     & \bf 46.1\textcolor{blue}{(2.10)} & \bf 45.5\textcolor{blue}{(1.80)} & \bf 44.3\textcolor{blue}{(0.40)} \\
\bottomrule
\end{tabular}}
\vspace{-3mm}
\caption{\small Performance comparison (Avg.) between baselines and our method on MMDU\cite{MMDU}.}
\label{tab:MMDU}
\vspace{-3mm}
\end{table}
\begin{figure}[!t]\small
    \centering
    \begin{minipage}[b]{0.525\linewidth}
        \centering        
        \includegraphics[width=\linewidth]{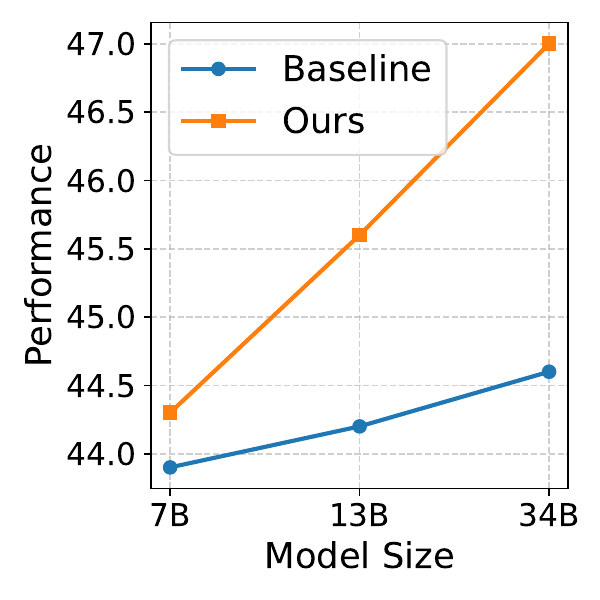}
        \vspace{-9mm}
        \caption{\small Performance (Avg.) across different sizes of LLaVA-Next on MMDU.}
        \label{fig:size_MMDU}
    \end{minipage}
    \hspace{0.02\linewidth}
    \begin{minipage}[b]{0.42\linewidth}\small
        \centering
        \setlength{\tabcolsep}{8.5pt}
        \begin{tabular}{lc}
        \toprule
        \textbf{Method} & \textbf{Avg.} \\
        \midrule
        Baseline & 43.9 \\
        SpAtten  & 43.5 \\
        LTP      & 44.0 \\
        \rowcolor{gray!15}Ours & \bf 44.3 \\
        \bottomrule
        \end{tabular}
        \vspace{-2mm}
        \captionof{table}{\small Performance comparison of our method, SpAtten \cite{SpAtten}, and LTP \cite{LTP} on MMDU with LLaVA-Next.}
        \label{tab:comparison_MMDU}
    \end{minipage}
\vspace{-5mm}
\end{figure}
The set $C_{ww}$ encompasses all token pairs except those already included in $C_{vr}$, $C_{vc}$, $C_{rt}$, $C_{ct}$, or $C_{vt}$. Specifically, $S_{vr}$ represents the flow from the visual part to preserved tokens, and $S_{vc}$ to pruned tokens. $S_{rt}$ and $S_{ct}$ capture the flows from preserved and pruned tokens to the target position, respectively. $S_{vt}$ denotes the direct flow from the visual part to the target, while $S_{ww}$ quantifies flow among all tokens, excluding those covered by the above flow. 

Figure~\ref{fig:flow} shows that pruning enhances the flow from visual inputs to preserved tokens ($S_{vr}$) and the target ($S_{vt}$), while reducing flow to pruned tokens ($S_{vc}$). These changes validate \textbf{Proposition 1} by demonstrating that pruning stabilizes and concentrates attention on important tokens, reducing entropy. They also support \textbf{Proposition 2}, as the increased $S_{vr}$ and $S_{vt}$ highlight the model’s shift towards greater dependency on visual inputs. More Analysis can be found in Appendix~\ref{app:FutherFlow}.

\section{Experiments}

\subsection{Experimental Setting}
To validate the effectiveness of our proposed method, we conduct experiments using the long-context dataset collected as mentioned in Sec.~\ref{sec:setting}, which is based on the SVIT dataset~\citep{SVIT}. It comprises 15,295 samples, divided into five groups with 0.5k, 1k, 1.5k, 2k, and 2.5k samples, respectively. The LVLMs with 7B parameters used in our experiments include DeepSeek-VL, LLaVA, LLaVA-Next, ViP-LLaVA, LLaVA-Next-Video, and Video-LLaMA2. The token pruning is applied to the first 10 layers, with a pruning rate of 0.3. For evaluation, we utilize greedy decoding and conduct model inference on NVIDIA A100 GPUs.

\subsection{Results}
In Table~\ref{tab:result}, we present the performance of different LVLMs across various context lengths, comparing the baseline methods with our approach. Our method demonstrates consistent improvements across all models, particularly for those struggling with long-context reasoning, such as DeepSeek-VL and ViP-LLaVA, where significant performance gains are observed. However, for models like LLaVA-Next and LLaVA-Next-Video, which already perform well at baseline, the improvements are relatively minor. 
To further analyze the effectiveness of our approach in such scenarios, we conducted additional experiments on the MMDU benchmark, as shown in Table~\ref{tab:MMDU}. These results reveal that our method provides consistent yet modest gains, especially for highly optimized models, highlighting its capability to refine performance even when baseline results are already strong. Additionally, due to token pruning, our method significantly improves inference efficiency, reducing inference time by approximately half compared to the original setup.
Moreover, Figure~\ref{fig:size_MMDU} shows that as the LLaVA-Next model size increases, the performance gap between our method and the baseline widens, demonstrating the scalability and effectiveness of our approach on larger models.
In Table~\ref{tab:comparison_MMDU}, our method achieves the highest average performance (44.3) on the MMDU benchmark with LLaVA-Next, outperforming both SpAtten~\cite{SpAtten} (43.5) and LTP~\cite{LTP} (44.0). This demonstrates our method's superior ability to refine performance and handle contextual information more effectively than these alternatives. Details of MMDU dataset and more results can be found in Appendix~\ref{app:MMDU}.

\subsection{Analysis on Token Pruning}
\paragraph{Impact of Token Pruning Method.}
\begin{figure}[t]
    \centering
    \includegraphics[width=\linewidth]{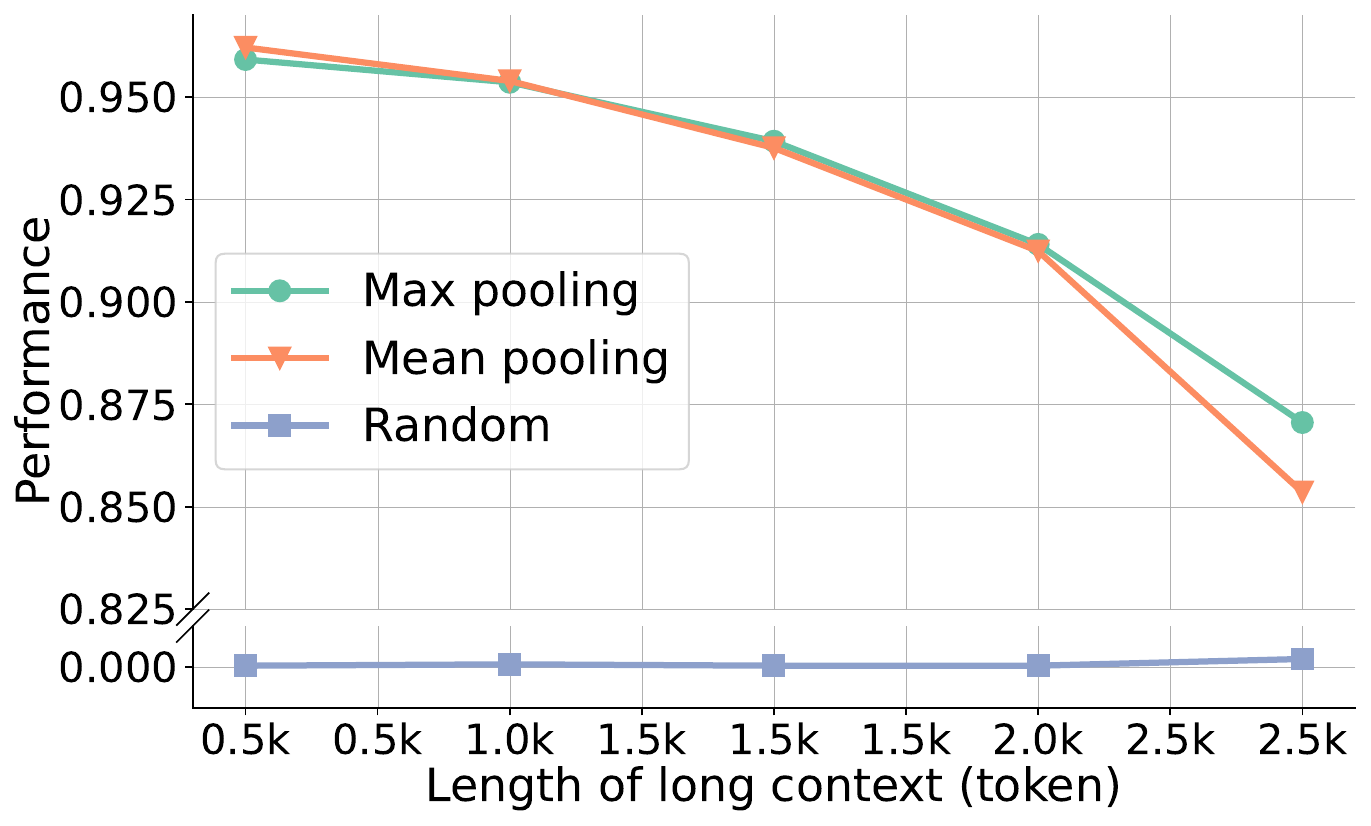}
    \vspace{-8mm}
    \caption{\small Comparison of results obtained using various token pruning methods. The random pruning approach ensures the final question token remains intact.}
    \label{fig:pruningmethod}
    \vspace{-4mm}
\end{figure}
Figure~\ref{fig:pruningmethod} shows the performance comparison of different token pruning methods: ``Max Pooling'', ``Mean Pooling'', and ``Random''. Both ``Max Pooling'' and ``Mean Pooling'' maintain high accuracy as the sequence length decreases, supporting our claim in \textbf{Proposition 1} that pruning low-importance tokens is crucial. These methods leverage multi-head attention to preserve key tokens with higher attention weights, ensuring minimal performance loss. In contrast, ``Random'' leads to a complete loss in accuracy, highlighting the importance of intelligently selecting tokens for pruning to maintain model performance. More analysis on ``Differences in Token Pruning Layers'', ``Impact of Token Pruning layer Numbers'', and ``Impact of Token Pruning Rate'' can be found in Appendix~\ref{app:layers}, Appendix~\ref{app:layer}, and Appendix~\ref{app:rate}, respectively.

\paragraph{Pruned Token Distribution.}
\begin{figure}[t]
    \centering
    \includegraphics[width=\linewidth]{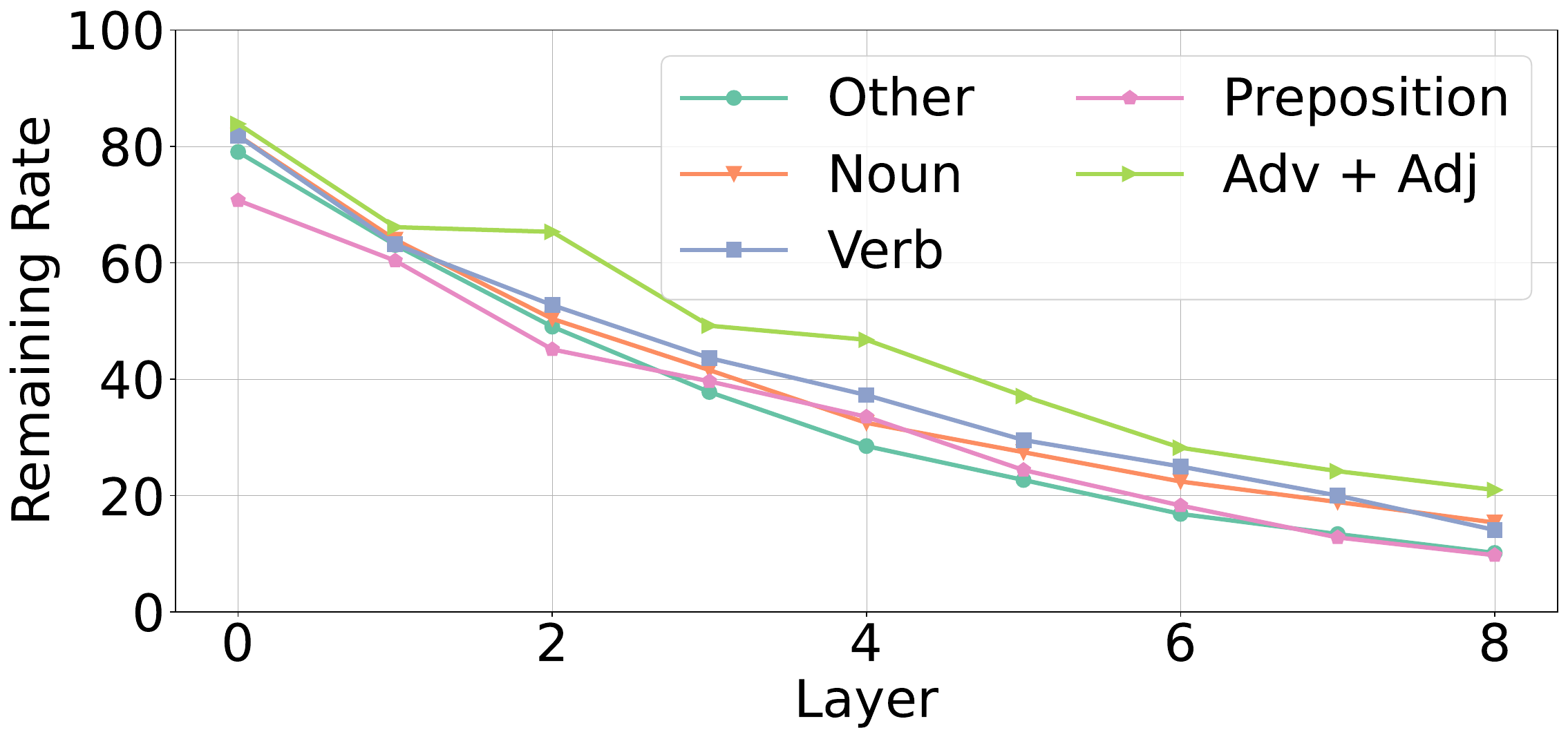}
    \vspace{-8mm}
    \caption{\small The remaining tokens after pruning across different layers of the model.}
    \label{fig:prunedtoken}
    \vspace{-4mm}
\end{figure}
Figure~\ref{fig:prunedtoken} illustrates the distribution of token types retained after pruning across model layers. While the overall proportion of token types remains fairly uniform, parts of speech show differing retention patterns. 
% Nouns and verbs consistently dominate the retained tokens, whereas prepositions and "other" tokens vary more across layers.
In contrast, Figure~\ref{fig:pruningmethod} highlights the severe performance drop caused by random pruning, which, despite similar pruning ratios, fails to retain critical information. Our method, as evidenced in Figure~\ref{fig:prunedtoken}, effectively identifies and removes less important tokens, preserving key ones and maintaining model performance even with significant pruning. The detailed analysis can be found in the Appendix~\ref{app:distribution}. More analysis on ``Token Pruning Scaling Laws'', ``Pruned Token Position'', and ``Case Study'' can be found in Appendix~\ref{app:ScalingLaws}, Appendix~\ref{app:Position}, and Appendix~\ref{app:Case}, respectively.

\section{Conclusion}
In this study, we investigate the challenges faced by LVLMs in long-context reasoning. Our empirical analysis demonstrated that LVLMs experience a significant decline in visual dependency as the context length increases. 
We also identified that 
% a higher proportion of target objects within images helps stabilize visual dependency and that 
cross-modal interactions are more prominent in the shallow layers of LVLMs, shifting towards predominantly textual interactions in deeper layers. 
To mitigate the performance degradation, we proposed a novel context pruning method that strategically removes less critical textual information. 
Our approach was rigorously validated through extensive experiments, demonstrating its effectiveness across various LVLMs.
% Additional experiments confirmed the robustness of different token pruning strategies and elucidated the relationship between pruning rates and context length.

\section*{Limitations}
While this study provides significant insights into the performance of LVLMs in long-context reasoning, there is still room for further research. The extensive computational resources required for our analysis limited the number of LVLMs we were able to evaluate. In addition, our investigation into Token Pruning Scaling Laws was conducted as a preliminary exploration due to resource constraints. Although we identified a relationship between context length and Token Pruning rate, more comprehensive studies are needed to elucidate this connection fully and to establish robust scaling laws.

\bibliography{custom}
\clearpage
\appendix
\section{Related Work}\label{app:related}
\subsection{Large Vision-Language Models}
Recent advancements in LVLMs have focused on enhancing cross-modal understanding by integrating visual and textual inputs for complex tasks~\citep{VisionLLM,MiniGPTv2,ShareGPT4V,Look}. For instance, LLaVA~\citep{llava} leverages GPT-4~\citep{GPT4} to generate multimodal instruction-following data, improving performance in tasks like Science QA and multimodal reasoning. Expanding on this, models such as SVIT~\citep{SVIT} and LLaVA-NeXT~\citep{LLaVANeXT} utilize larger datasets and improve capabilities in OCR and abstract reasoning. Approaches like DeepSeek-VL~\citep{DeepSeekVL} address real-world challenges by handling noisy and ambiguous inputs, while ViP-LLaVA~\citep{ViPLLaVA} enhances robustness in processing diverse visual prompts. Additionally, some methods such as SURf’s self-refinement framework improve the accuracy of information retrieval~\citep{surf}, and VICL~\citep{VICL} enables few-shot learning through inference on visual captioning. \citet{LLaVANeXTVideo} extends to zero-shot video understanding using AnyRes, allowing models trained on images to generalize to videos, while Video-LLaMA2 ~\citep{VideoLLaMA2} enhances spatial-temporal modeling by integrating visual and audio inputs. These developments highlight LVLMs' trends in scaling data, improving reasoning, and increasing robustness.

\subsection{Long-Context Reasoning}
Handling long contexts is a key challenge for LLM. Traditional position encoding techniques often struggle with long-range dependencies or varying-length sequences. To address this, methods like Contextual Position Encoding (CoPE)~\citep{Contextual} dynamically adjust positional encodings based on context, improving performance on tasks requiring long-range dependencies. Training-free techniques, such as contextual sliding windows~\citep{LongContext}, extend the context length of pre-trained models without retraining, reducing computational overhead while maintaining sequence-processing capabilities. Parallel Context Windows (PCW)~\citep{Parallel} further enhances efficiency by splitting sequences into chunks for parallel processing.
However, models tend to underutilize information from earlier inputs, favoring the middle or end~\citep{LostintheMiddle}. Some methods like ThoT~\citep{ThoT} address this by improving reasoning across fragmented contexts, while LLMLingua~\citep{LLMLingua} compresses prompt representations, striking a balance between efficiency and context retention.
Similarly, SpAtten~\citep{SpAtten} proposes an efficient sparse attention mechanism that leverages cascade token and head pruning to dynamically reduce computational overhead, significantly improving the model's performance. LTMP~\citep{LTP} introduces learned thresholds for token merging and pruning in Vision Transformers, achieving high token reduction rates with minimal accuracy loss by learning optimal trade-offs between merging and pruning across layers. These approaches showcase promising strategies to manage computational complexity while maintaining model performance across diverse tasks.

\section{Detail Analysis of Consistency Between Vision and Language Priors}\label{app:consistency}
To investigate the prediction consistency between vision and language priors, we analyze the disjoint portions of predictions with and without visual priors.
Figure \ref{fig:consistency} presents the average performance of six models on different context lengths, comparing the accuracy of models provided visual priors (VL correct) and those only using language priors (Text correct). The proportion of samples that were correctly answered only when provided with an image generally decreases as the context length increases. It indicates that while visual dependency is more pronounced in shorter contexts, its impact diminishes as the model is exposed to more textual information. This suggests that the model increasingly relies on language reasoning under longer contexts. The proportion of samples only correctly answered without an image remains near zero across all context lengths. 
The results show that samples correctly answered using only language information remained to be correct even after adding visual information.
It suggests that the models have effectively leveraged language priors to make accurate predictions, regardless of the availability of visual cues. As the context length increases, models rely more on the rich semantic information from text, reducing visual dependency.

\begin{table*}[!t]\small
\centering
\begin{tabular}{lcccccccc}
\toprule
\textbf{Model} & \textbf{Method} & \textbf{C} & \textbf{R} & \textbf{VP} & \textbf{LC} & \textbf{AA} & \textbf{IRU} & \textbf{Avg.} \\
\midrule
\multirow{2}{*}{LLaVA-Next-Video \cite{LLaVANeXTVideo}} & Baseline & 38.5 & 39.4 & 42.1 & 57.6 & 45.8 & 40.7 & 44.0 \\
                                  & \cellcolor{gray!15} Ours & \cellcolor{gray!15}40.9 & \cellcolor{gray!15}42.0 & \cellcolor{gray!15}42.8 & \cellcolor{gray!15}58.0 & \cellcolor{gray!15}49.5 & \cellcolor{gray!15}43.5 & \cellcolor{gray!15}46.1 \\
\midrule
\multirow{2}{*}{Video-LLaMA2 \cite{VideoLLaMA2}}     & Baseline & 37.7 & 39.3 & 41.6 & 57.2 & 45.9 & 40.2 & 43.7 \\
                                  & \cellcolor{gray!15} Ours & \cellcolor{gray!15}39.7 & \cellcolor{gray!15}39.4 & \cellcolor{gray!15}43.3 & \cellcolor{gray!15}58.4 & \cellcolor{gray!15}49.6 & \cellcolor{gray!15}42.8 & \cellcolor{gray!15}45.5 \\
\midrule
\multirow{2}{*}{LLaVA-Next \cite{LLaVANeXT}}       & Baseline & 38.2 & 39.3 & 41.7 & 57.6 & 45.8 & 40.5 & 43.9 \\
                                  & \cellcolor{gray!15} Ours & \cellcolor{gray!15}38.8 & \cellcolor{gray!15}39.8 & \cellcolor{gray!15}42.0 & \cellcolor{gray!15}57.9 & \cellcolor{gray!15}46.1 & \cellcolor{gray!15}41.1 & \cellcolor{gray!15}44.3 \\
\bottomrule
\end{tabular}
\caption{\small Performance comparison between baselines and our method on MMDU.}
\label{tab:MMDU_all}
\end{table*}
\begin{table*}[!t]\small
\centering
\begin{tabular}{lcccccccc}
\toprule
\textbf{Size} & \textbf{Method} & \textbf{C} & \textbf{R} & \textbf{VP} & \textbf{LC} & \textbf{AA} & \textbf{IRU} & \textbf{Avg.} \\
\midrule
\multirow{2}{*}{7B}  & Baseline & 38.2 & 39.3 & 41.7 & 57.6 & 45.8 & 40.5 & 43.9 \\
                     & \cellcolor{gray!15} Ours & \cellcolor{gray!15}38.8 & \cellcolor{gray!15}39.8 & \cellcolor{gray!15}42.0 & \cellcolor{gray!15}57.9 & \cellcolor{gray!15}46.1 & \cellcolor{gray!15}41.1 & \cellcolor{gray!15}44.3 \\
\midrule
\multirow{2}{*}{13B} & Baseline & 38.4 & 39.7 & 41.8 & 58.0 & 46.5 & 40.8 & 44.2 \\
                     & \cellcolor{gray!15} Ours & \cellcolor{gray!15}40.8 & \cellcolor{gray!15}41.2 & \cellcolor{gray!15}43.6 & \cellcolor{gray!15}58.3 & \cellcolor{gray!15}46.9 & \cellcolor{gray!15}42.8 & \cellcolor{gray!15}45.6 \\
\midrule
\multirow{2}{*}{34B} & Baseline & 38.7 & 40.1 & 42.5 & 58.3 & 46.9 & 40.8 & 44.6 \\
                     & \cellcolor{gray!15} Ours & \cellcolor{gray!15}40.7 & \cellcolor{gray!15}43.3 & \cellcolor{gray!15}45.9 & \cellcolor{gray!15}60.4 & \cellcolor{gray!15}50.3 & \cellcolor{gray!15}41.2 & \cellcolor{gray!15}47.0 \\
\bottomrule
\end{tabular}
\caption{\small Performance comparison across different model sizes (7B, 13B, 34B) of LLaVA-Next on MMDU.}
\label{tab:size_MMDU_all}
\end{table*}
\begin{table*}[!t]\small
\centering
\begin{tabular}{lccccccc}
\toprule
\textbf{Method} & \textbf{C} & \textbf{R} & \textbf{VP} & \textbf{LC} & \textbf{AA} & \textbf{IRU} & \textbf{Avg.} \\
\midrule
Baseline & 38.2 & 39.3 & 41.7 & 57.6 & 45.8 & 40.5 & 43.9 \\
SpAtten \cite{SpAtten}  & 38.0 & 38.5 & 41.5 & 57.2 & 45.7 & 40.4 & 43.5 \\
LTP \cite{LTP}      & 38.3 & 38.9 & 42.1 & 57.7 & 46.0 & 40.8 & 44.0 \\
\cellcolor{gray!15}Ours & \cellcolor{gray!15}38.8 & \cellcolor{gray!15}39.8 & \cellcolor{gray!15}42.0 & \cellcolor{gray!15}57.9 & \cellcolor{gray!15}46.1 & \cellcolor{gray!15}41.1 & \cellcolor{gray!15}44.3 \\
\bottomrule
\end{tabular}
\caption{\small Performance comparison of our method with SpAtten and LTP on MMDU, using the LLaVA-Next.}
\label{tab:comparison_MMDU_all}
\end{table*}
\section{Further Analysis of Information Flow}\label{app:FutherFlow}
\begin{figure}[t]
    \centering
    \includegraphics[width=1\linewidth]{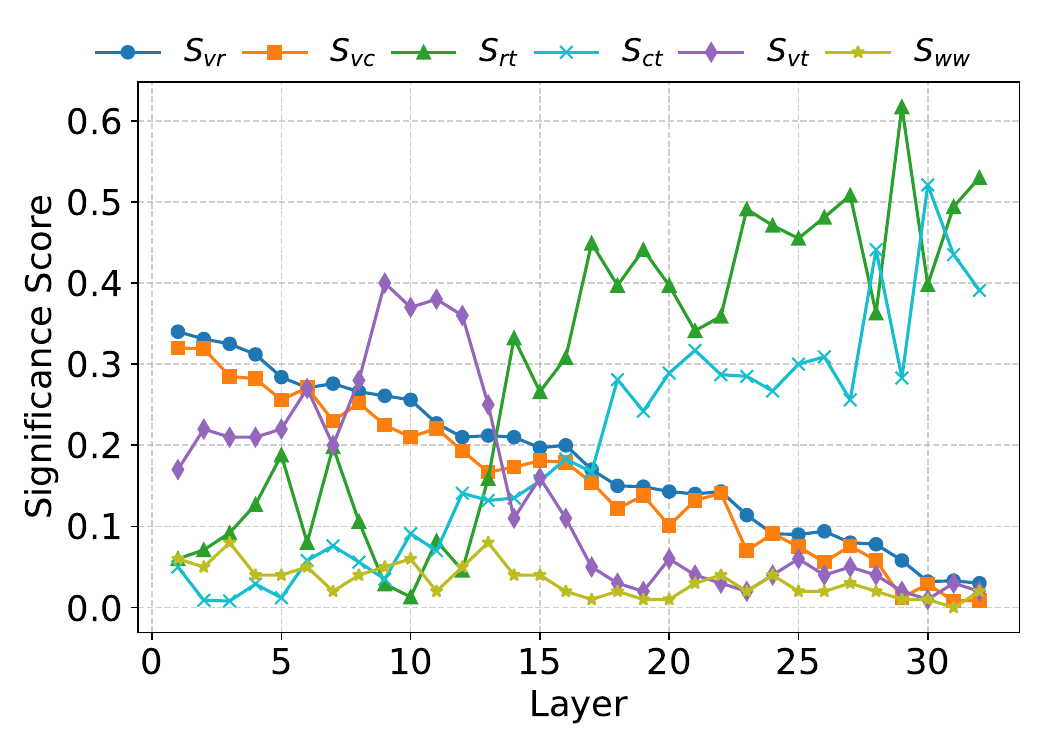}
    \caption{Information flow results from baseline.}
    \label{fig:flow_pre}
\end{figure}

Figure~\ref{fig:flow_pre} and Figure~\ref{fig:flow}
In this section, we extend the analysis of information flow by comparing the baseline results (Figure~\ref{fig:flow_pre}) with the pruned model (Figure~\ref{fig:flow}). 
From Figure~\ref{fig:flow_pre}, it is evident that the baseline model distributes information flow more evenly across preserved tokens ($S_{vr}$), pruned tokens ($S_{vc}$), and the target ($S_{vt}$). However, the significance scores of $S_{vr}$ and $S_{vt}$ remain relatively lower in comparison to the pruned model. This indicates that, in the absence of pruning, the model's attention is less focused, leading to higher entropy and a less stable flow of information.
In contrast, the pruned results in Figure~\ref{fig:flow} show a noticeable increase in $S_{vr}$ and $S_{vt}$, alongside a significant drop in $S_{vc}$. This shift reflects the model's improved ability to concentrate attention on essential tokens, aligning with \textbf{Proposition 1}, which states that pruning reduces entropy by stabilizing and concentrating attention. Moreover, the consistent rise in $S_{vr}$ and $S_{vt}$ across deeper layers highlights the model's increasing reliance on visual inputs for decision-making, supporting \textbf{Proposition 2}.

\section{More Results on MMDU}\label{app:MMDU}
This section presents detailed experimental results on the MMDU dataset, which is designed to evaluate multimodal capabilities across six metrics: Creativity (C), Richness (R), Visual Perception (VP), Logical Coherence (LC), Answer Accuracy (AA), and Image Relationship Understanding (IRU). Each of these metrics provides a comprehensive assessment of a model's ability to generate coherent and meaningful responses in tasks involving multimodal data.
The MMDU dataset consists of multimodal tasks where each sample we used includes two images, testing a model's ability to process and relate visual inputs effectively. Performance evaluation follows the methodology outlined by \citet{MMDU}, utilizing a set of standardized evaluation prompts. All models were assessed using GPT-4o, which serves as the scoring framework to ensure consistency and robustness in the evaluation process. Tables~\ref{tab:MMDU_all}, \ref{tab:size_MMDU_all}, and \ref{tab:comparison_MMDU_all} summarize the results of our method compared to baselines and prior approaches. Key findings are highlighted below:  

\paragraph{Comparison with Baselines.}  
As shown in Table~\ref{tab:MMDU_all}, our approach consistently outperforms baseline methods across all six metrics for different models, including LLaVA-Next-Video~\cite{LLaVANeXTVideo}, Video-LLaMA2~\cite{VideoLLaMA2}, and LLaVA-Next~\cite{LLaVANeXT}. Notably, the average performance improvement ranges from 0.4 to 2.1 points, demonstrating the efficacy of our proposed enhancements.  

\paragraph{Effect of Model Size.}  
Table~\ref{tab:size_MMDU_all} presents a detailed breakdown of performance for different model sizes (7B, 13B, and 34B). Larger models, such as the 34B version, show significant improvements, particularly in \textbf{Visual Perception (VP)} and \textbf{Logical Coherence (LC)}, where our method achieves up to 2.4 points of improvement over the baseline. These results underscore the scalability of our approach as model size increases.  

\paragraph{Comparison with Prior Methods.}  
In Table~\ref{tab:comparison_MMDU_all}, we compare our method with SpAtten~\cite{SpAtten} and LTP~\cite{LTP} using the LLaVA-Next model. Our method achieves the highest scores across all metrics, with an average performance gain of 0.3 points compared to LTP, which is the closest competitor.

\section{Differences in Token Pruning Layers.}\label{app:layers}
\begin{figure}[t]
    \centering
    \includegraphics[width=\linewidth]{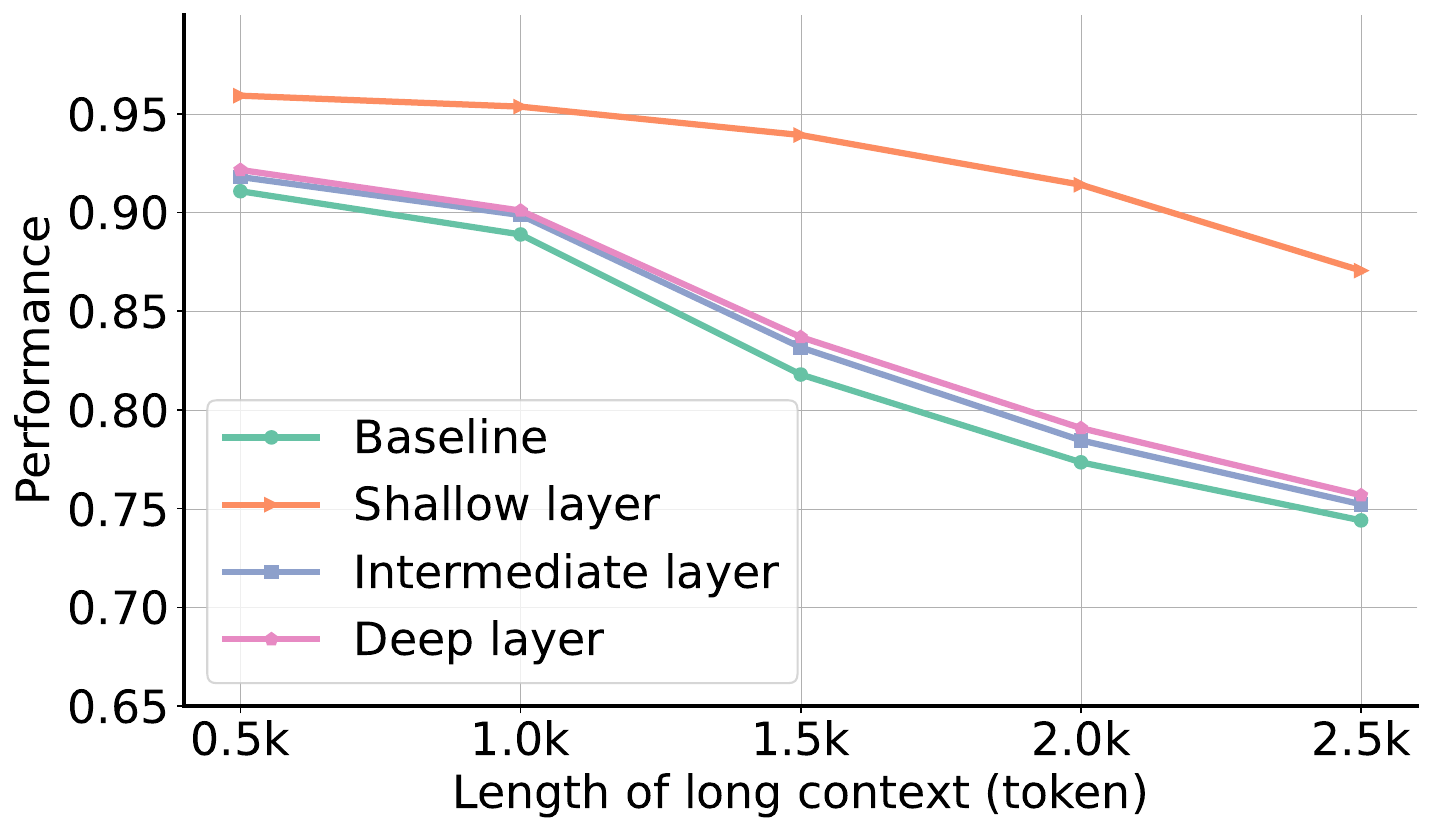}
    \caption{\small Results with different token pruning layers.}
    \label{fig:pruninglayers}
\end{figure}

To investigate the impact of token pruning at different layers, we analyze the performance of our method when pruning tokens at the shallow, intermediate, and deep layers for varying input lengths. As shown in Figure~\ref{fig:pruninglayers}, results demonstrate that token pruning in the shallow layers consistently achieves better performance across all tested input lengths, while pruning in later layers (i.e., intermediate and deep layers) results in a noticeable performance drop. The performance remains relatively unchanged compared to the baseline when pruning is conducted in the later layers. These findings confirm our observations from Figure~\ref{fig:attention}, where we noted that attention weights between text and vision interactions are more significant in the shallow layers of LVLMs. This indicates that the shallow layers are more effective at identifying tokens that are less relevant to the visual information, making them ideal for filtering. By pruning tokens in these layers, we can preserve the most important information for cross-modal understanding. 

\section{Impact of Token Pruning layer Numbers.}\label{app:layer}
\begin{figure}[t]
\centering
    \includegraphics[width=\linewidth]{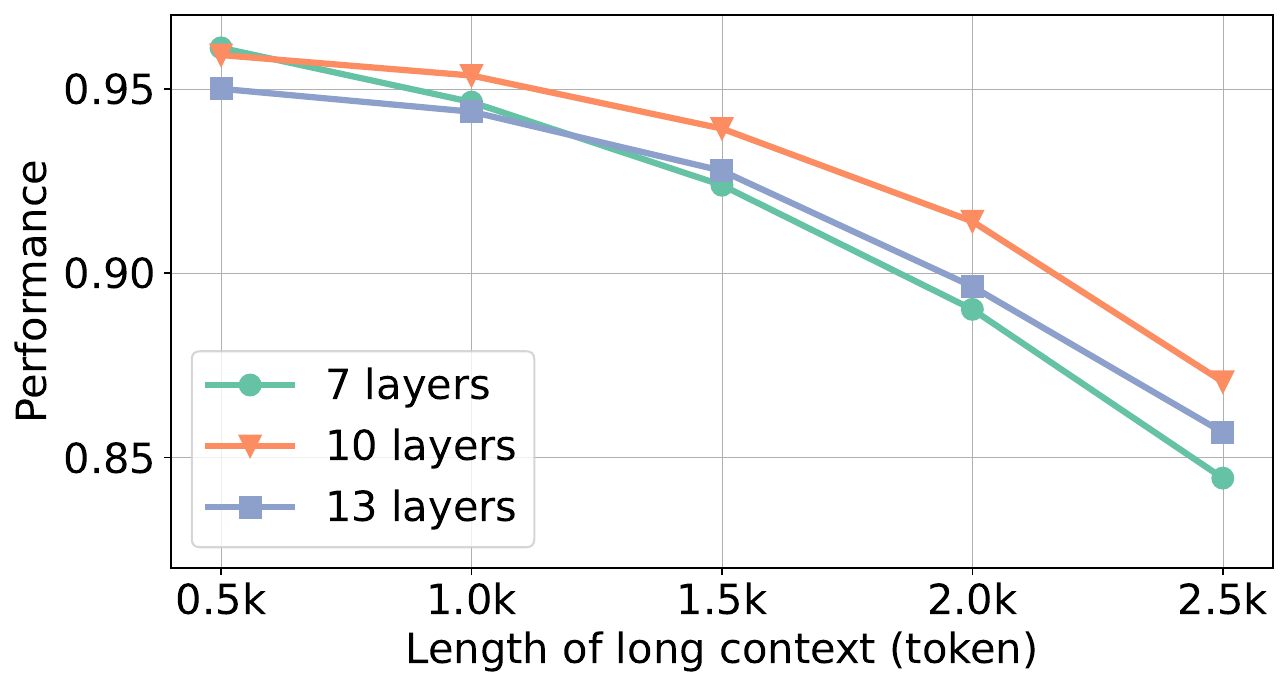}
    \caption{\small Performance of token pruning with different layer numbers.}
    \label{fig:pruninglayernumber}
\end{figure}
Figure~\ref{fig:pruninglayernumber} shows the impact of applying Token Pruning at different layer depths (7, 10, and 13 layers) across varying sequence lengths. The results indicate that deeper layers (10 and 13) generally achieve higher accuracy, particularly for longer sequences. While the accuracy difference is minimal for shorter sequences (0.5k and 1.0k tokens), it becomes more pronounced as the sequence length increases, with 10 layers outperforming other layers significantly at 2.5k tokens. 

\section{Impact of Token Pruning Rate.}\label{app:rate}
\begin{figure}[t]
    \centering
    \includegraphics[width=\linewidth]{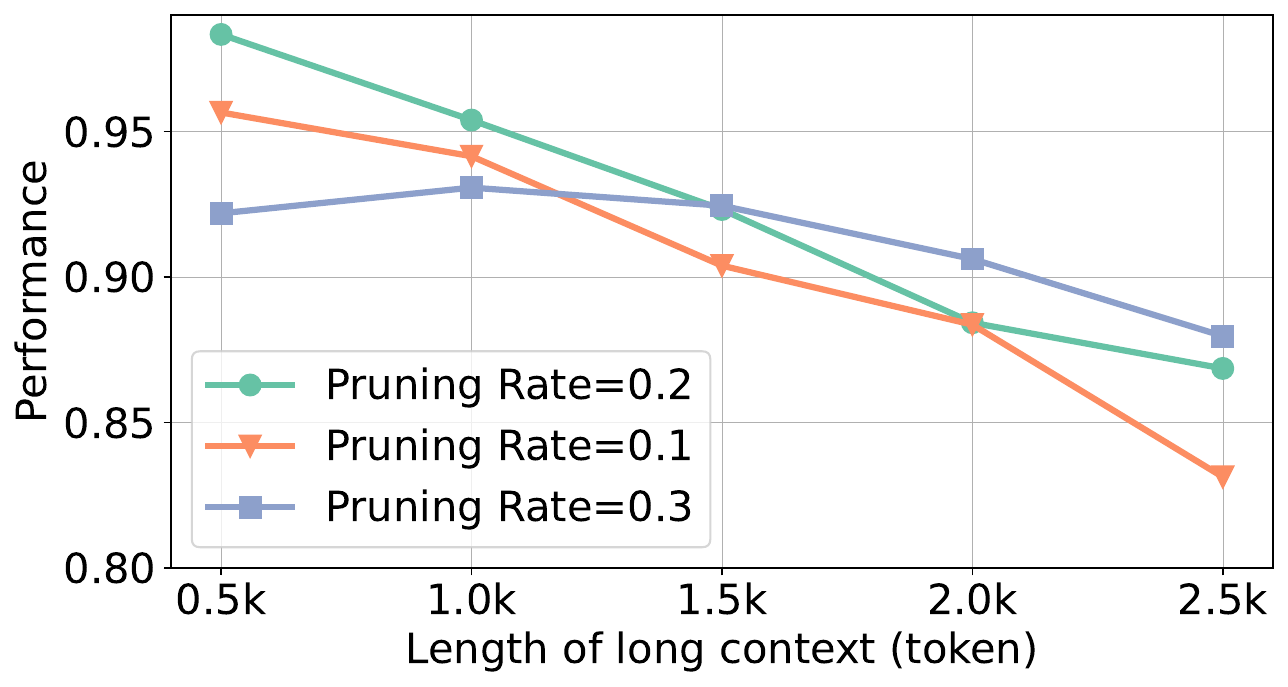}
    \caption{\small Results with different token pruning rates.}
    \label{fig:pruningrate}
\end{figure}
To analyze the impact of token pruning rates on performance, we evaluate our method with different pruning rates (0.1, 0.2, and 0.3) across varying input lengths (0.5k to 2.5k tokens), as shown in Figure \ref{fig:pruningrate}. The results indicate that for shorter sequences (0.5k and 1.0k), performance remains high across all pruning rates, with only a slight drop as the pruning rate increases. However, for longer sequences (1.5k to 2.5k), higher pruning rates lead to an improvement in performance, especially at the 0.3 pruning rate compared to 0.1 and 0.2. It demonstrates that a fixed pruning rate may not be suitable for all input lengths. Therefore, a dynamic pruning strategy that adjusts the rate based on the input length could better preserve accuracy across different sequence sizes.

\section{Detailed Analysis on Pruned Token Distribution}\label{app:distribution}
Figure~\ref{fig:prunedtoken} shows the distribution of token types remaining after pruning at each layer of the model. It can be observed that although the proportion of remaining token types is relatively uniform across layers, different parts of speech exhibit varying degrees of retention as pruning progresses. For instance, nouns and verbs consistently make up a significant portion of the retained tokens, while prepositions and tokens in the ``other'' category show greater variability across layers. This result is in stark contrast to the performance degradation observed in Figure~\ref{fig:pruningmethod}, where random token pruning leads to significantly worse performance. Although both approaches maintain a relatively uniform token pruning ratio across layers, the random pruning method fails to preserve critical information, as evidenced by its poor performance. However, results from Figure~\ref{fig:prunedtoken} demonstrate that our pruning method effectively identifies and removes less important tokens, retaining those that are more crucial for model performance. This highlights the strength of our approach in accurately selecting the tokens to prune, thus preserving model performance even under significant pruning.

\section{Token Pruning Scaling Laws}\label{app:ScalingLaws}
\begin{figure}[t]
    \centering
    \includegraphics[width=\linewidth]{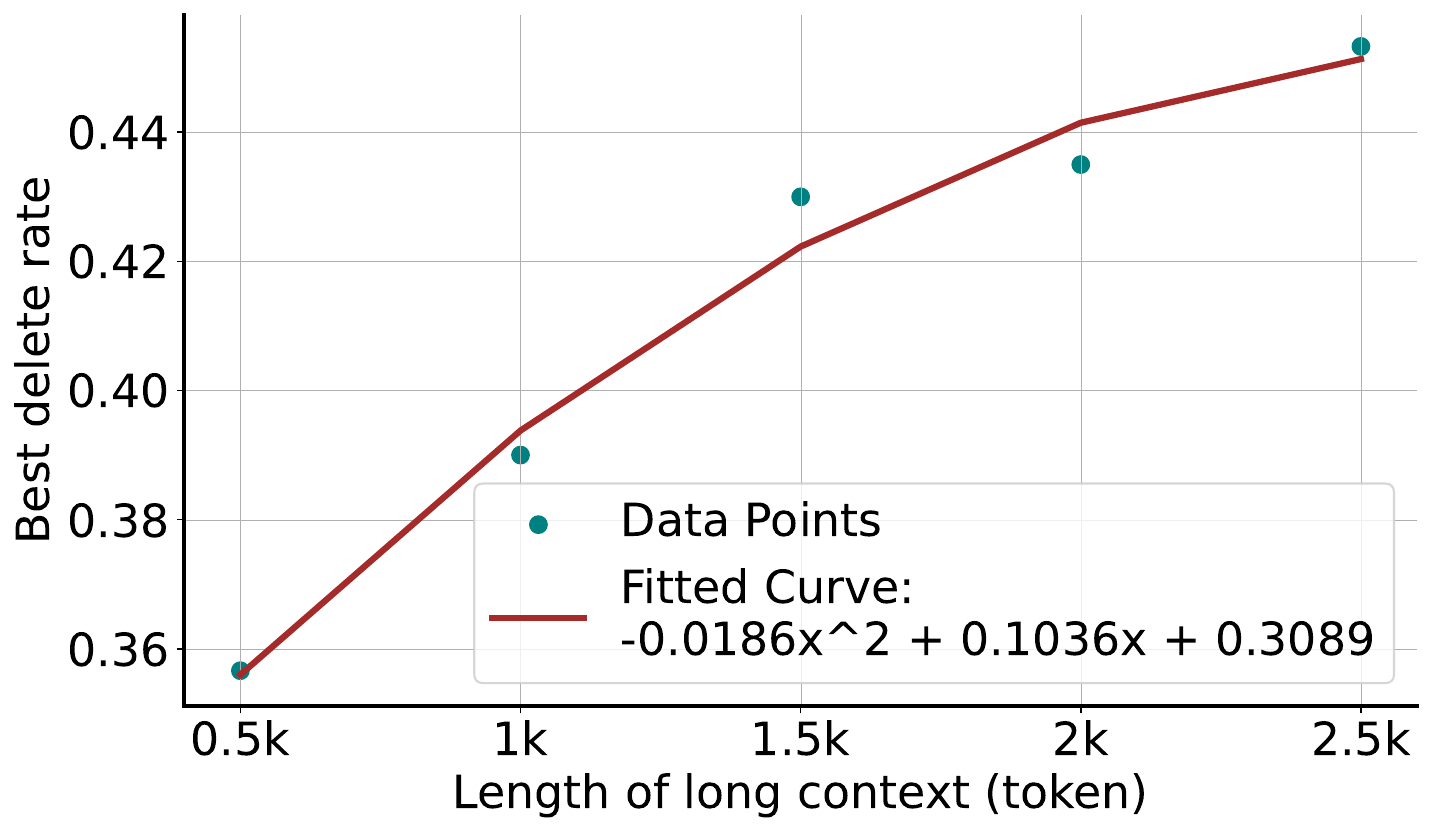}
    \caption{\small Token Pruning Scaling Laws.}
    \label{fig:scalinglaws}
\end{figure}
\begin{figure}[t]
    \centering
    \includegraphics[width=\linewidth]{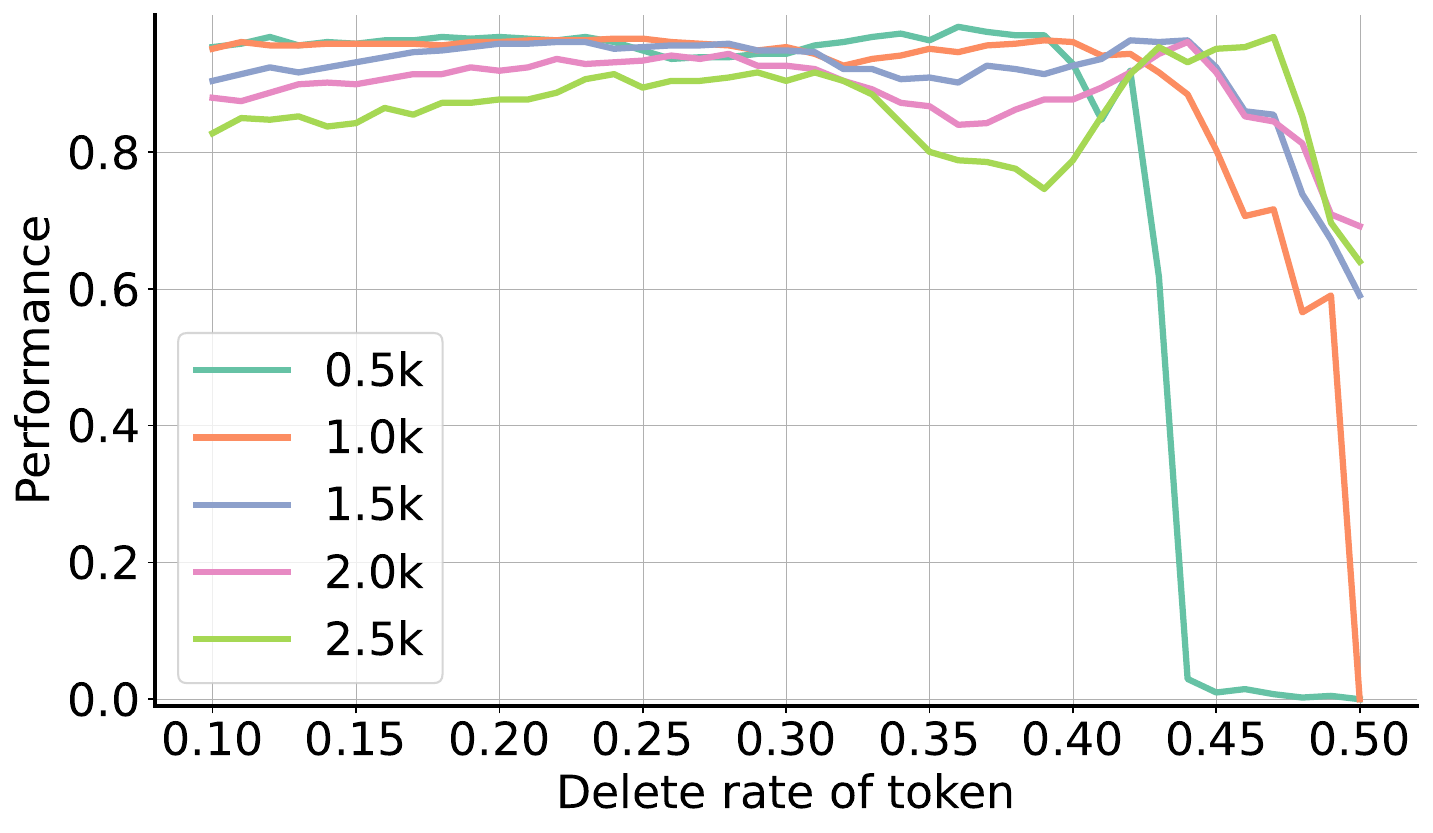}
    \caption{\small Performance of LVLMs with different token pruning rates and context lengths.}
    \label{fig:ratesandlengths}
\end{figure}
Figure~\ref{fig:scalinglaws} shows the relationship between the pruning rate and performance in long-context reasoning. The ``best pruning rate'' represents averaging the top three adjacent performance points from testing various pruning rates. 
The performance for different pruning rates and lengths is shown in Figure~\ref{fig:ratesandlengths}.
The plotted points correspond to the best pruning rates at different context lengths, while the red curve represents a fitted quadratic regression line, capturing the trend in the relationship between these variables. From the curve and data points, we observe a clear upward trend, indicating that as the token delete rate increases (i.e., as the context length extends), the best delete rate also rises. It demonstrates that for longer contexts, more aggressive pruning may be advantageous in preserving optimal performance.

\section{Pruned Token Position}\label{app:Position}
\begin{figure}[t]
    \centering
    \includegraphics[width=\linewidth]{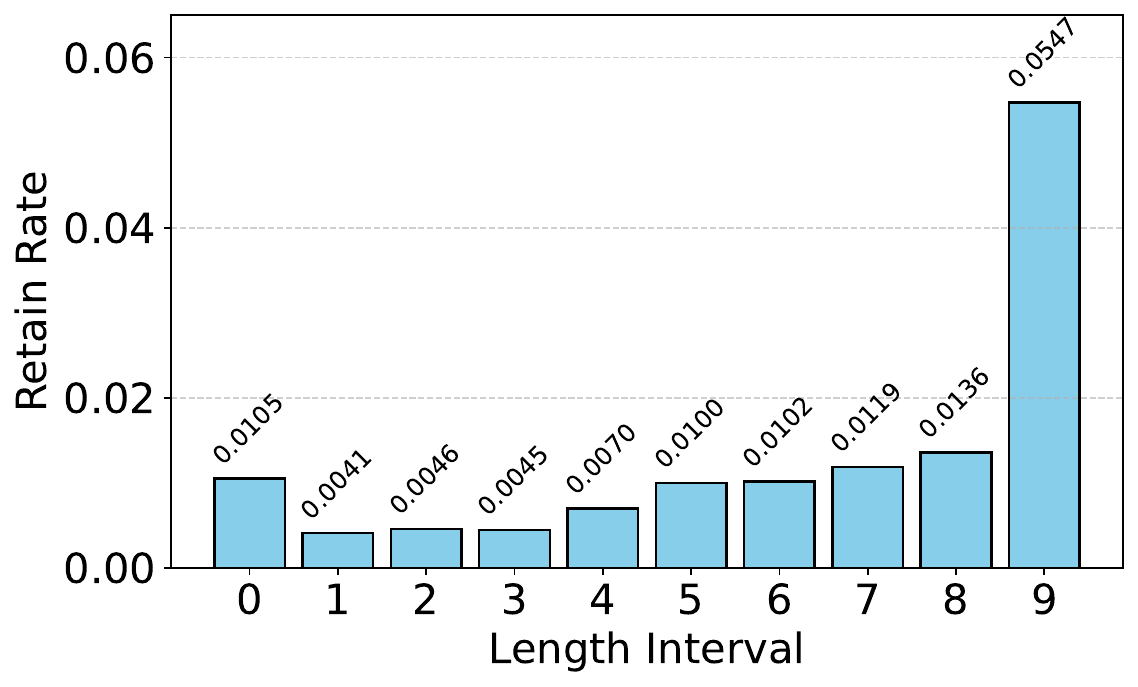}
    \caption{\small Token retain rates across different length intervals for pruned tokens using our method.}
    \label{fig:retain_rates}
\end{figure}
Figure~\ref{fig:retain_rates} shows the retained rate of tokens across different length intervals within the context. Our method demonstrates relatively uniform pruning across the earlier intervals (0–8), while tokens in the final interval (9) are retained at a significantly higher rate. Because the final interval often contains the question tokens, which are essential for generating accurate responses. The even pruning of intervals ensures efficient token usage, preserving essential information to improve model performance.

\section{Case Study}\label{app:Case}
\begin{figure*}[!t]
    \centering
    \includegraphics[width=\linewidth]{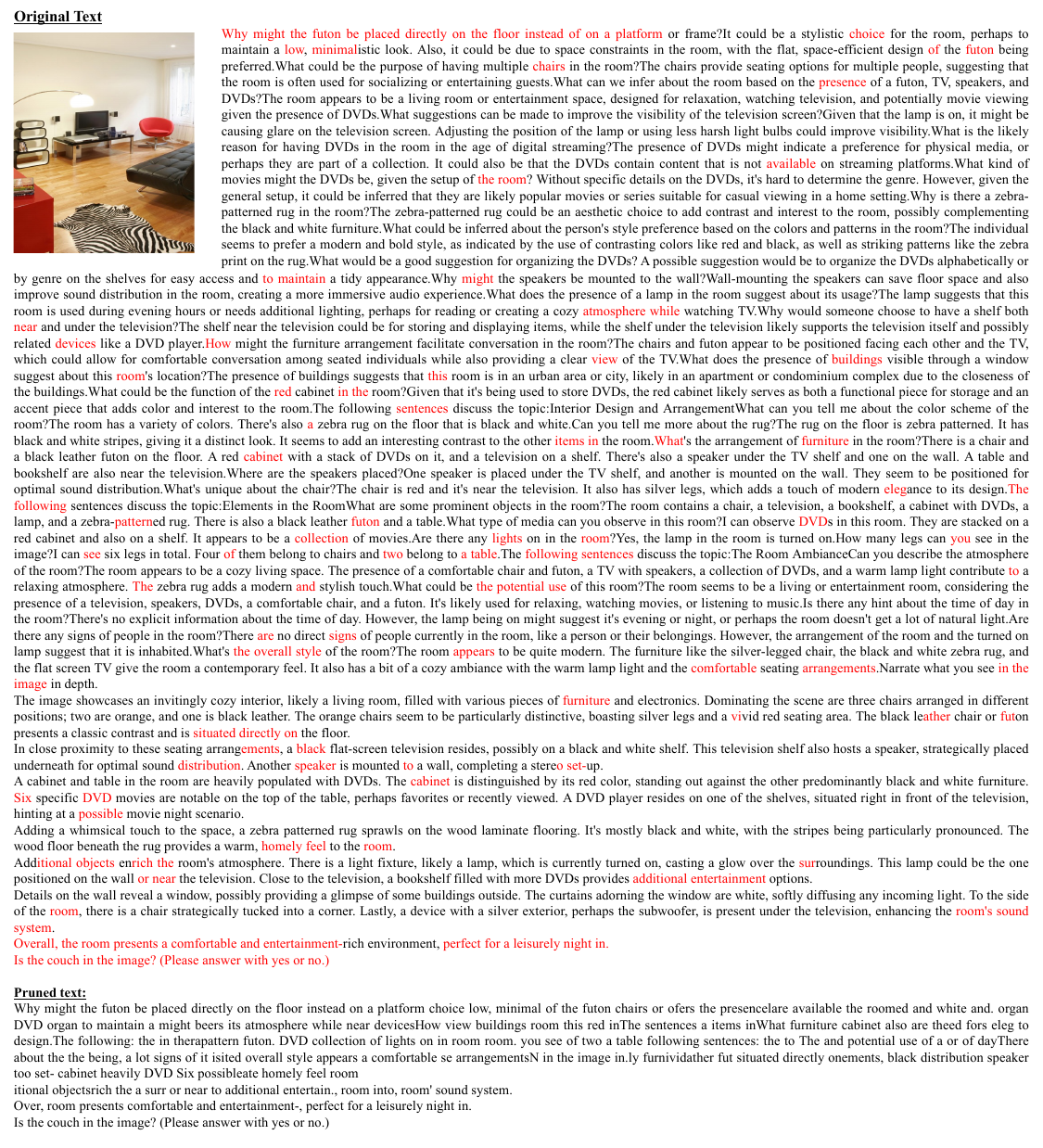}
    \caption{\small Case of token pruning using our method, with the red parts indicating the retained tokens.}
    \label{fig:case}
\end{figure*}

Our method effectively removes most irrelevant tokens, retaining only those related to the context. As shown in Fig.~\ref{fig:case}, the baseline method consistently fails in long-context reasoning tasks with contexts longer than 0.5k tokens, whereas our approach correctly answers across various context lengths.

\end{document}